\documentclass{egpubl}
\usepackage{egsr2026}

\SpecialIssuePaper         

\CGFStandardLicense

\usepackage[T1]{fontenc}
\usepackage{dfadobe}
\usepackage[normalem]{ulem}

\usepackage{cite}  
\BibtexOrBiblatex
\electronicVersion
\PrintedOrElectronic
\ifpdf \usepackage[pdftex]{graphicx} \pdfcompresslevel=9
\else \usepackage[dvips]{graphicx} \fi

\usepackage{egweblnk}

\usepackage{booktabs} 
\usepackage{xspace}
\usepackage{soul}

\usepackage{color}
\usepackage{graphicx}
\usepackage{amsmath}
\usepackage{multirow}
\usepackage{array}
\usepackage[capitalize,nameinlink]{cleveref}

\usepackage[english]{babel}
\usepackage[utf8]{inputenc}
\usepackage{algpseudocode}
\usepackage{overpic}
\usepackage{afterpage}
\usepackage{amssymb}
\usepackage{orcidlink}
\usepackage{hyperref}

\title{ResEdit: Residual embeddings for precise generative image editing}

\author[Baykal et al.]{
    \parbox{\textwidth}{%
        \centering%
        Canberk Baykal$^{1,2}$\orcidlink{0000-0002-0249-5858} \quad
        Valentin Deschaintre$^{1}$\orcidlink{0000-0002-6219-3747} \quad
        Yannick Hold-Geoffroy$^{1}$\orcidlink{0000-0002-1060-6941} \quad
        Michael Fischer$^{1}$\orcidlink{0000-0002-2610-4831} \quad
        Anna Frühstück$^{1}$\orcidlink{0000-0002-3870-4850} \\[1mm]
        Cengiz Öztireli$^{2}$\orcidlink{0000-0002-4700-2236} \quad
        Iliyan Georgiev$^{1}$\orcidlink{0000-0002-9655-2138} \quad
    }
    \vspace{1mm}\\
    $^1$Adobe Research \quad $^2$University of Cambridge
}

\usepackage{booktabs} 
\usepackage[capitalize,nameinlink]{cleveref}
\usepackage{soul}
\usepackage{xfrac}
\usepackage{xcolor}

\usepackage[most]{tcolorbox}
\tcbset{on line, boxsep=0pt, left=1pt, right=1pt, top=1pt, bottom=1pt, colback=yellow, colframe=yellow, sharp corners}
\usepackage{soul} 

\DeclareGraphicsExtensions{.png,.jpg,.pdf,.ai,.psd}
\newcommand{\Paragraph}[1]{\paragraph*{#1.}}

\newcommand{\model}{f}
\newcommand{\params}{\theta}
\newcommand{\paramsprobe}{\phi}
\newcommand{\probe}{g_\paramsprobe}
\newcommand{\network}{v_\params}
\newcommand{\target}{v^*}
\newcommand{\image}{I}
\newcommand{\latent}{z}

\newcommand{\condition}{C}
\newcommand{\residual}{R}
\newcommand{\eps}{\varepsilon}

\newcommand{\lossrecon}{\mathcal{L}_\text{recon}}
\newcommand{\lossprobe}{\mathcal{L}_\text{probe}}
\newcommand{\schedule}{\sigma}
\newcommand{\encoder}{\mathcal{E}}

\newcommand{\rgbx}{RGB$\rightarrow$X\xspace}
\newcommand{\xrgb}{X$\rightarrow$RGB\xspace}
\newcommand{\rgbxx}{RGB$\leftrightarrow$X\xspace}
\newcommand{\rgbxrgb}{RGB$\rightarrow$X$\rightarrow$RGB\xspace}

\usepackage[ruled]{algorithm2e} 

\SetAlFnt{\small}
\SetAlCapFnt{\small}
\SetAlCapNameFnt{\small}
\SetAlCapHSkip{0pt}

\DeclareGraphicsExtensions{.png,.jpg,.pdf,.ai,.psd}
\begin{document}


\teaser{
    \vspace{-6mm}
    \includegraphics[width=\linewidth]{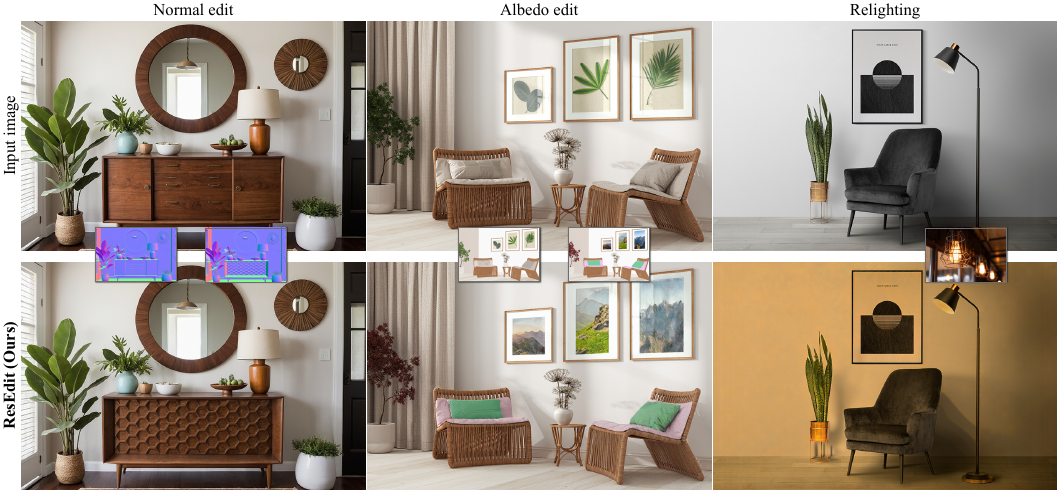}%
    \vspace{-2.5mm}%
    \caption{
        Our framework enables high-fidelity generative editing by isolating image identity into a learned residual image embedding. Unlike traditional inversion methods that struggle with ``baked-in'' condition features, ResEdit explicitly separates identity from the physical condition, facilitating streamlined intrinsic-space manipulation of geometry and material, as well as reference-based relighting. By shifting the burden of reconstruction from the noise latent to this dedicated residual channel, we achieve a superior balance of identity preservation and responsive editability without necessitating model surgery.
    }
    \label{fig:teaser}
}

\maketitle


\begin{abstract}
Conditional diffusion image generators can be repurposed for editing through inversion, without the need for large-scale paired fine-tuning data. However, producing high-quality, targeted edits while maintaining image identity and global consistency remains challenging, as weakly conditioned inversion often embeds conflicting image features into the noise. We demonstrate that incorporating a residual image encoding as additional conditioning enables both improved identity preservation and better editability. We optimize this residual encoding to provide a strong conditioning signal for reconstruction, thereby reducing the reliance on inversion and susceptibility to its aforementioned pitfalls. To ensure this residual does not interfere with desired edits, we incorporate a gradient reversal-based optimization strategy that disentangles the residual from the edited condition. We illustrate our method's ability to produce high-fidelity results across precise intrinsic-based editing and relighting, and show proof-of-concept text-guided manipulation. 
Project page: \href{https://johnberg1.github.io/resedit/}{johnberg1.github.io/resedit}
\end{abstract}

\section{Introduction}

Image editing is central to modern creative workflows, yet it has traditionally demanded substantial technical expertise and meticulous effort. Recently, generative diffusion models \cite{ho2020denoising, dhariwal2021diffusion, rombach2022high, esser2024scaling} have matured into a powerful foundation for a new editing paradigm, enabling complex manipulations through intuitive interactions. In this work, we tackle the problem of identity-preserving image editing, with a primary focus on intrinsic-space manipulation \cite{luo2024intrinsic, kocsis2024intrinsic, zeng2024rgbx, lyu2025intrinsicedit}, which facilitates precise, physically-grounded control over an image's underlying geometry, material, and illumination (see \cref{fig:teaser}).

Two basic approaches exist for generative image editing. The first established approach is to repurpose a \emph{conditional generation} model, by inverting the input image with respect to a conditioning signal, modifying the condition, and then generating an edited result from the inverted noise \cite{song2021denoising,mokady2023null,samuel2024lightning}. This approach is very general, but it introduces a fundamental tension between achieving a plausible edit and preserving the input's identity \cite{ju2024pnp,nguyen2025swiftedit}. The second approach is to train a dedicated \emph{editing} model that directly maps an input image and an edit specification (e.g., a text prompt) to an (edited) output image \cite{brooks2023instructpix2pix}. Recent industrial-scale implementations have shown remarkable abilities \cite{gpt4oimage,FLUX1context,google2025nanobanana}, though with control limited to text-based prompting. Moreover, editing models need large-scale paired-image training data, which is difficult and expensive to obtain \cite{wang2025seededit}. Since paired data generation typically relies on editing-repurposed generation models \cite{liu2025step1xedit}, advancing the paired-data-free approach remains a prerequisite for scaling editing capabilities.

While diffusion generators can produce high-fidelity images from scratch based on an input condition, they often struggle with precision, fidelity, and plausibility when used for editing via inversion, frequently failing to preserve unedited regions. Recent works try to mitigate this by optimizing a prompt condition during inversion \cite{dong2023prompt,mokady2023null,miyake2025negative} or manipulating internal cross-attention maps \cite{hertz2023prompt,tumanyan2023plug,cao2023masactrl}. Yet, these approaches do not explicitly address the inherent entanglement of spatial structure and semantic appearance within the latent space. This entanglement causes localized edits to either ``leak'' into global changes or result in a rigid reconstruction that resists significant modifications. More recently, approaches such as \rgbxx~\cite{zeng2024rgbx} and IntrinsicEdit~\cite{lyu2025intrinsicedit} pioneered coupling these ideas with intrinsic image decomposition \cite{barrow1978recovering}, allowing artists to directly edit intrinsic channels (e.g., albedo, lighting). IntrinsicEdit, for instance, optimizes a prompt-embedding condition to reduce the incorporation of image features into the noise during inversion and to mitigate conflicts between edited and unedited channels. Despite this, identity leakage can still occur during inversion, limiting editing precision and degrading image fidelity beyond the edit. 

In this paper, we present ResEdit, a method that achieves precise image editing with a simplified user experience. Our core contribution is the introduction of a \emph{residual image embedding}, which improves current methods in two ways: (1)~\emph{Improved identity preservation:} Our residual embedding provides complementary, side-channel information that acts as a strong signal for image content, enabling high-fidelity reconstruction, even without inversion. (2)~\emph{Better disentanglement:} We employ gradient reversal \cite{ganin2015gradient} to penalize information sharing between the residual and the input intrinsic channel. The input and residual conditions thus act as complementary signals, achieving both precise reconstruction and high sensitivity to edits, significantly reducing the reliance on inversion and susceptibility to noise entanglement artifacts. 

We validate our method experimentally, demonstrating improved editing quality compared to existing methods across various intrinsic-decomposition-based editing tasks. Our versatile residual representation is compatible with other side-channel information; for example, we integrate the recently proposed UniLight embedding \cite{zhang2025unilight} to create an easy-to-use relighting pipeline. Furthermore, we present proof-of-concept results showing that our residual strategy can also benefit text-based image editing, providing initial evidence that the same formulation applies beyond intrinsic-editing scenarios. Our residual formulation streamlines the editing process, opening the door to fast editing and improved user experience.

In summary, we propose a general generative image editing framework that utilizes a residual image embedding to preserve identity and enable editability without requiring structural surgery to the diffusion model. We demonstrate its versatility across several tasks, including intrinsic-space editing, relighting, and proof-of-concept text-based manipulation. Unlike prior intrinsic editing approaches that require processing auxiliary channels and resolving inter-channel conflicts, ours significantly streamlines the user experience by operating solely on the specific condition targeted for editing. We also show a prototype feed-forward model that replaces the iterative residual optimization with a single inference pass, significantly reducing latency and paving the way for real-time, precise generative image editing. We will release our code and models.

\section{Related work}
\label{sec:related_work}

\Paragraph{Generative image editing}

Diffusion models have turned image editing into a process of conditional denoising. The goal is to retain the image's layout and identity while changing its semantic content. Early work demonstrated adding noise to a (possibly crudely manipulated) image and then denoising it~\cite{meng2022sdedit}, with a trade-off between realism, control, and identity preservation. With the rise of large text-to-image models, editing has shifted toward cross-attention manipulation~\cite{hertz2023prompt, cao2023masactrl} and instruction following~\cite{brooks2023instructpix2pix}. Recent work has advanced training-free editability~\cite{zhu2025training} toward more challenging transformations, such as geometric edits. In-context editing models like FLUX.1 Kontext~\cite{FLUX1context} take the original image and the edit instruction together to ensure edits stay on-target and preserve identity. This capability is enabled by large-scale training datasets containing before/after-edit pairs, which are created by editing-repurposed generation models. More recent general-purpose image editing models, including Nano Banana 2~\cite{google2026nanobanana2}, FLUX.2~\cite{flux-2-2025}, and Qwen-Image-Edit~\cite{wu2025qwenimagetechnicalreport} provide strong prompt and reference-based editing capabilities. These models target broad instruction-following image editing, whereas our focus is precise intrinsic-space manipulation.

\Paragraph{Diffusion inversion}

Editing real images using a diffusion model usually starts by recovering a noise trajectory that reconstructs the input. The classic DDIM inversion~\cite{song2021denoising} is fast but suffers from discretization error accumulation, which is exacerbated under strong classifier-free guidance (CFG) \cite{ho2022classifier}. Exact inversion~\cite{hong2024on} solves a fixed-point optimization problem at each diffusion step, improving reconstruction fidelity but at higher cost. GNRI \cite{samuel2024lightning} provides a significantly faster and more accurate alternative. Our method employs inversion, but exhibits low sensitivity to the precision of the latent trajectory.

\Paragraph{Prompt optimization for inversion/editing}

Another line of work on inversion involves optimizing conditioning rather than only the noise to mitigate identity drifts during editing. Textual inversion~\cite{gal2022image} learns tokens for a reusable visual concept. Null-text inversion~\cite{mokady2023null} optimizes the unconditional prompt embedding used in CFG, keeping the conditional prompt fixed; negative-prompt inversion \cite{miyake2025negative} offers an optimization-free alternative. Prompt-tuning inversion~\cite{dong2023prompt} learns image-specific conditional embeddings while P2L~\cite{chung2024prompt} optimizes text embeddings on-the-fly during inversion. DATE \cite{na2025diffusion} updates prompt embeddings during sampling via timestep-adaptive text embedding updates. Motivated by these methods, we propose a disentangled residual image embedding that encodes details to facilitate both reconstruction and editing. Residual-based refinement has also been explored for GAN inversion~\cite{alaluf2021restyle}, and recent diffusion-based methods optimize inversion for downstream editing~\cite{xu2026invert}. Unlike prior work, our residual tokens are optimized per image as a complementary identity channel and explicitly discouraged from encoding the editable condition.

\Paragraph{Adversarial disentanglement}

Disentangled representations in generative modeling include GAN-based formulations, which learn factorization by maximizing mutual information between latents and the generated image \cite{chen2016infogan} or by enforcing the latent dimensions to be independent \cite{kim2018disentangling}. Closer to our setup, some approaches explicitly enforce invariance through an adversarial game: an additional module learns to predict the factor from the representation, and the representation itself is learned so the factor becomes more difficult to predict~\cite{ganin2017domain, edwards2016censoring}. Fader networks~\cite{lample2017fader} apply this idea to controllable manipulation by making the latents invariant to chosen attributes. In the diffusion setting, recent methods suggest that text-based diffusion models can support attribute disentanglement in their conditioning space~\cite{wu2023uncovering}. Motivated by these ideas, we adversarially disentangle our learned residual embedding from the edited input condition.

\Paragraph{Intrinsic decomposition \& editing}

Intrinsic image decomposition aims to factor an RGB image into physically meaningful components representing geometry, materials, and illumination. Recent methods treat this as a generative problem and use diffusion models to sample intrinsic modalities~\cite{kocsis2024intrinsic,luo2024intrinsic}. Careaga and Aksoy \cite{careaga2024colorful} relax the single-illuminant assumption and explicitly model colorful shading and specular residuals.

Paired with a corresponding re-rendering model, intrinsic decomposition enables precise editing by manipulating individual intrinsic channels. Relighting works emphasize lighting representations derived from intrinsic cues and explicitly target preservation of intrinsic properties~\cite{kocsis2024lightit}. ZeroComp~\cite{zhang2025zerocomp} supports inserting synthetic objects into a real image through its intrinsic channels, resolving lighting via a diffusion model. Diffusion-based intrinsic estimation has also been applied to scene-level appearance editing~\cite{kocsis2024intrinsic}. IntrinsicControlNet~\cite{lu2025intrinsiccontrolnet} conditions image generation on intrinsic maps using ControlNet-style~\cite{zhang2023adding} geometry and material branches.

Closest to our work, \rgbxx~\cite{zeng2024rgbx} pairs diffusion models to enable intrinsic decomposition (\rgbx) as well as controllable image generation (\xrgb) and editing. A naive \rgbxrgb approach using this pipeline suffers from identity drifts and editing difficulty due to intrinsic-channel entanglement. IntrinsicEdit~\cite{lyu2025intrinsicedit} addresses these issues via exact \xrgb inversion and prompt tuning. Our approach combines token optimization and adversarial learning into a per-image residual identity channel that complements the user-editable condition, streamlining the editing process and delivering higher-fidelity results.

\begin{figure*}
    \centering
    \includegraphics[width=\linewidth]{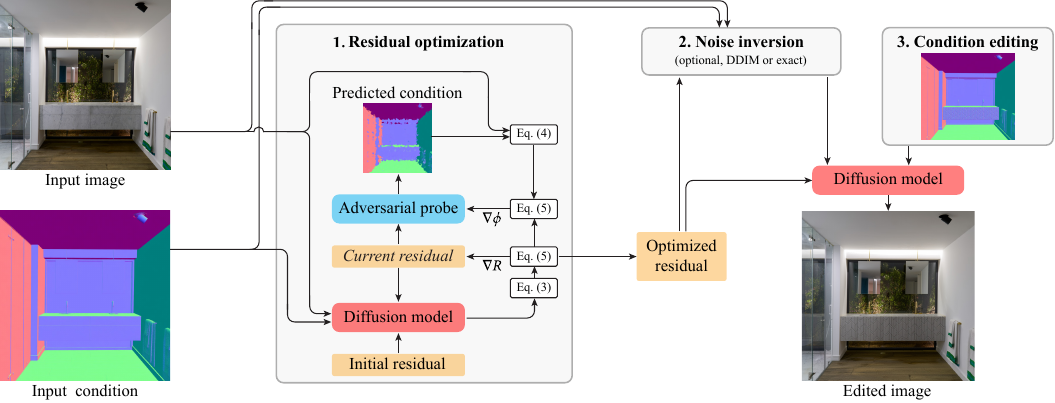}%
    \vspace{-1.5mm}
    \caption{
        \textbf{Method overview.}
        Our framework enables high-fidelity editing through a three-stage process: (1)~Given an input image and its corresponding condition (e.g., albedo), we optimize a residual embedding to capture the input identity. An adversarial loss is employed to disentangle the residual from the condition, encoding only the missing identity. (2)~The input image is inverted relative to the joint conditioning signal to obtain a noise latent. Because the residual narrows the generative distribution, this inversion is more stable and often optional. (3)~The edited condition is combined with the fixed residual and noise to produce the final edited image, which preserves the source identity while faithfully reflecting the conditional manipulation.
    }
    \label{fig:overview}
\end{figure*}

\section{Method}
\label{sec:method}

We aim to adapt a pre-trained conditional diffusion model for high-fidelity image editing by manipulating its input condition. Our main use case is intrinsic-space editing, where the condition represents an intrinsic channel (e.g., albedo); however, we will intentionally keep our exposition application-agnostic. Our objective is to generate a new image that reflects the change in the condition while faithfully preserving the source's identity. The standard approach is to utilize diffusion inversion to map the source image into a noise latent that, when denoised under the original condition, reconstructs the input. One could expect that resynthesizing from this latent with a modified condition should produce the desired edit. However, in practice, this approach suffers from a fundamental entanglement between the conditioning signal and the inverted noise, necessitating a trade-off between editability and identity preservation \cite{mokady2023null,hong2024on,ju2024pnp,kulikov2025flowedit} or the use of additional application- and model-specific manipulation mechanisms to realize the edit \cite{hertz2023prompt}.

To overcome this entanglement and enable responsive and high-fidelity editing, we propose a framework that shifts the burden of identity preservation from the noise manifold to a new, learnable \emph{residual image embedding}. This residual serves as a new conditioning signal that is optimized to capture image identity while remaining disentangled from the input condition. The two complementary conditions anchor the model to the image, thereby also reducing the reliance on inversion and, thus, the susceptibility to its characteristic artifacts.

In the following, we formalize our problem setting, describe the construction of our residual embedding, and lay out our full editing pipeline. An illustration of that pipeline is shown in \cref{fig:overview}.

\subsection{Problem setup}

We consider a diffusion model $\model$ that, given a condition $\condition$ and a normally distributed noise $\latent \sim \mathcal{N}(\mathbf{0},\mathbf{1})$, generates an image $\image$ consistent with the condition:
\begin{equation}
    \image = \model(\latent \mid \condition).
\end{equation}
We assume that the model is a deterministic function of its arguments, e.g., inference follows DDIM~\cite{song2021denoising} or flow matching \cite{Lipman2023FlowMatching,Liu2023RectifiedFlow}. Inverting the model means finding the noise that reconstructs a given image under a given condition, i.e., $\latent = \model^{-1}(\image \mid \condition)$.

\begin{figure*}
    \centering
    \includegraphics[width=\linewidth]{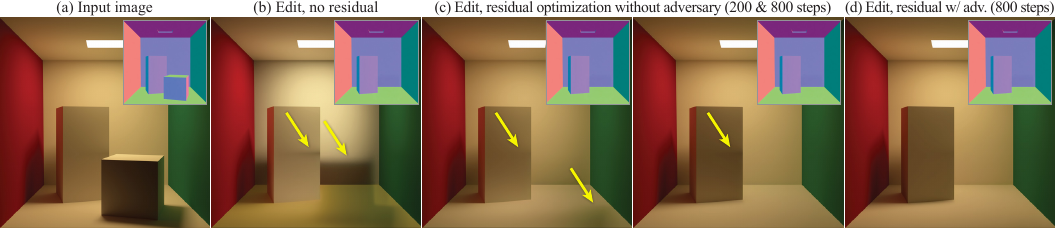}%
    \vspace{-1.5mm}
    \caption{
        \textbf{Residual effect.}
        We show the effect of our residual on object removal via normal-map manipulation.
        (a)~Input image and normal map.
        (b)~Without a residual, source geometry and appearance are ``baked'' into the (exactly) inverted noise latent, causing the object to persist after manipulation.
        (c)~Optimizing our residual for reconstruction alone offers more flexibility than raw inversion, but entanglement with the input condition still manifests as ghosting artifacts.
        (d)~Our adversarial optimization improves representation disentanglement, enabling clean object removal with faithful identity preservation in~unedited~regions.
    }
    \label{fig:residual_spectrum}
\end{figure*}

While inversion can achieve accurate reconstruction \cite{song2021denoising,li2024source,hong2024on}, it does not inherently facilitate plausible editing by manipulating the condition \cite{lyu2025intrinsicedit}. By design, the condition provides only a partial specification of the scene; hence the inverted noise captures all remaining details not explained by the condition, such as lighting effects or geometry, and also encodes how the model generates those details in the image \emph{given the condition}, as illustrated in \cref{fig:residual_spectrum}b. This can cause the model to either resist subsequent modifications or suffer from significant identity drift when applying the edit. Furthermore, even minor condition modifications can trigger drastic, unintended global changes in the output \cite{hertz2023prompt, mahajan2024prompting}. This behavior is particularly pronounced in modern diffusion transformers, which employ joint multi-modal attention between text and image latents \cite{esser2024scaling}.

To mitigate the aforementioned issues, some existing methods optimize prompt embeddings during inversion to enforce editability via strong classifier-free guidance \cite{dong2023prompt, mokady2023null, miyake2025negative}. Others manipulate internal attention maps to decouple the inverted noise from the source condition \cite{hertz2023prompt, tumanyan2023plug, cao2023masactrl}. Ultimately, however, these approaches remain constrained by a fundamental and inherent trade-off between reconstruction fidelity and editability \cite{hou2024high, ouyang2025proedit} as they lack the degrees of freedom required to decouple source identity from the rigid structure of the inverted latent.

\subsection{Residual image embedding}

To address the trade-off between reconstruction and editability, we move away from the paradigm of relying solely on the latent noise to encode image identity. Instead, we conceptually extend the model conditioning with an additional embedding $\residual$:
\begin{equation}
    \image = \model(\latent \mid \condition, \residual).
\end{equation}
In practice, we instantiate $\residual$ as a set of optimizable tokens inserted into the model's text-token stream.

The model is now conditioned on two distinct signals: the original, interpretable condition $\condition$ and the new, learned \emph{residual image embedding} $\residual$. The goal is for them to work in tandem to provide strong guidance to the model and facilitate both image reconstruction and editing. For reconstruction, the residual $\residual$ should thus be optimized to encode any identity details of the source image $\image$ that are not represented by the original condition $\condition$. Crucially, to ensure editability, the residual should encode \emph{only} those missing details. In other words, we want $\residual$ to be disentangled from $\condition$, so that the information in $\residual$ will not conflict with any edits to $\condition$.

In this framework, the latent noise $\latent$ remains a generic stochastic scaffold, while the heavy lifting of identity preservation is shifted to the complementary conditions $\condition$ and $\residual$. We will show below that subsequent inversion can even become optional as the distribution of images parameterized by $\latent$ is effectively narrowed down by the stronger conditioning.

\subsection{Residual optimization}

Having defined the objectives of our residual embedding $\residual$, we need to inject it into the diffusion model and optimize it accordingly. Modern diffusion backbones provide strong text-conditioning mechanisms \cite{rombach2022high,esser2024scaling} and can be further fine-tuned with extra conditioning, e.g., for intrinsic-based neural rendering \cite{zeng2024rgbx}. We take such a frozen (potentially fine-tuned) model and expand its flexible token-based text conditioning with new tokens that will encode our residual. Text-token optimization has been successfully used in prior works for various tasks \cite{gal2022image,dong2023prompt,mokady2023null}.

Without loss of generality, we assume a flow-matching latent-diffusion model \cite{Lipman2023FlowMatching,Liu2023RectifiedFlow,esser2024scaling} implementing a neural network with parameters $\params$ that predicts a velocity field $\network(t, \latent_t, \condition)$ to denoise a noisy latent $\latent_t$ at time $t \in [0,1]$ back to a clean latent image representation $\latent_0$. To keep notation simple, we will denote the conditioning expansion by replacing the model's existing conditioning $\condition$ with $\{ \condition, \residual \}$. Here, $\condition$ remains the original, editable condition, while $\residual$ is an additional residual condition injected through the model's text-conditioning branch.

\Paragraph{Reconstruction}

To mitigate the noise entanglement pitfalls of inversion, we want the residual $\residual$ to provide a strong conditioning signal that enables high-fidelity reconstruction of the input image $\image$ given the input condition $\condition$, with any noise. To that end, we optimize $\residual$ using the model's reconstruction loss
\begin{equation}
    \label{eq:ddpm_loss}
    \lossrecon(\residual) = \mathbb{E}_{t, \eps}\,\Big\| \, \network\big(t, \latent_t, \{ \condition, \residual \} \big) - \target \Big\|^2.
\end{equation}
At each optimization iteration, we sample a random time step $t \in [0,1]$ and Gaussian noise $\eps \sim \mathcal{N}(\mathbf{0},\mathbf{1})$. The loss target is the velocity $\target = \eps - \latent_0$, where $\latent_0 = \encoder(\image)$ is the VAE-encoded latent representation of the input image. The noisy latent $\latent_t$ is obtained by applying the model's noising schedule $\schedule_t \in [0,1]$ to the input latent: $\latent_t = (1 - \schedule_t) \latent_0 + \schedule_t \eps$.

\Paragraph{Disentanglement}

To facilitate editing of the input condition, we want the residual to strictly complement it in reconstructing the image. The above optimization naturally achieves some disentanglement, as $\residual$ does not need to encode any image features already present in $\condition$. However, this behavior is not enforced explicitly and features can still leak through, pulling synthesis back toward the original image during editing, as shown in \cref{fig:residual_spectrum}c.

We mitigate this entanglement by adversarially discouraging $\residual$ from encoding the condition $\condition$. Specifically, we introduce a probe module $\probe$ whose task is to predict $\condition$ from $\residual$:
\begin{equation}
    \lossprobe(\paramsprobe, \residual) = \big\|\, \probe(\residual) - \condition \, \big\|_2^2.
    \label{eq:lprobe}
\end{equation}
We train the probe to minimize $\lossprobe$ while simultaneously training the residual $\residual$ to maximize that loss, to make it uninformative about $\condition$ but without sacrificing reconstruction. Our final optimization problem thus comprises the following two objectives:
\begin{equation}
    \min_{\residual} \big[\lossrecon(\residual) - \lambda\,\lossprobe(\paramsprobe, \residual) \big],
    \quad
    \min_{\paramsprobe}\ \lossprobe(\paramsprobe, \residual).
    \label{eq:minmax}
\end{equation}
We solve \cref{eq:minmax} per image using alternating updates. Each iteration performs one update to $\paramsprobe$ (with $\residual$ frozen) followed by one update to $\residual$ (with $\paramsprobe$ frozen). We begin with a warm-up phase in which both $\residual$ and $\paramsprobe$ are optimized, but the adversarial term is disabled.

\Paragraph{Discussion}

\begin{figure}[t]
    \includegraphics[width=\linewidth]{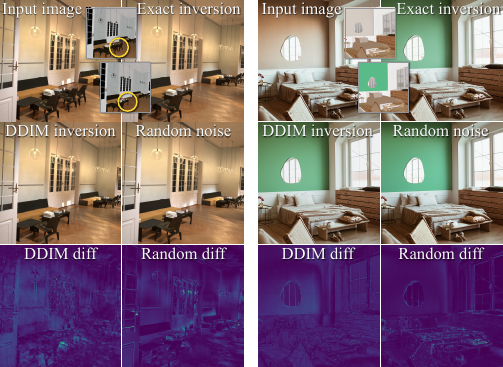}%
    \vspace{-1.5mm}
    \caption{%
        \textbf{Noise ablations.} The intrinsic condition and residual together enable high-quality edits with DDIM inversion and even with random noise, indicating that our residual effectively captures identity and narrows down the generative distribution.
    }
    \vspace{-3mm}
    \label{fig:noise_ablation}
\end{figure}

Note that our residual optimization is performed prior to any potential diffusion inversion, similarly to IntrinsicEdit \cite{lyu2025intrinsicedit}. In contrast, established editing methods \cite{dong2023prompt, mokady2023null} often perform prompt optimization \emph{after} inversion---at a point where noise entanglement has already manifested and is significantly harder to mitigate. By optimizing the residual before the noise is constrained by a specific trajectory, we gain the flexibility to capture the identity signal and decouple it from the source condition. Together, the input condition and the learned residual narrow the model's generative distribution, significantly reducing the sensitivity of the final output to the inversion process and mitigating the stiffness and identity drift characteristic of post-inversion optimization. In fact, we observe that a well-optimized residual can substantially reduce the need for precise inversion, enabling both high-fidelity reconstruction and responsive editing using random noise initialization, as shown in \cref{fig:residual_capacity,fig:noise_ablation}.

\subsection{Editing pipeline}

While the paper's primary focus is on intrinsic-space editing, our general pipeline, illustrated in \cref{fig:overview}, is application-agnostic. Given the input image and corresponding condition, we first optimize our residual embedding adversarially according to the objectives in \cref{eq:minmax} as described above. The residual and input conditions complement each other to anchor the model to the image. Diffusion inversion can then be performed w.r.t.\@ these conditions to pick up any remaining detail. Finally, the input condition is modified, and diffusion inference with the joint conditions and the noise yields an edited output image.

In the case of intrinsic-space editing, the condition targeted for editing is an intrinsic channel (e.g., albedo). Our pipeline consumes only one condition, in addition to the input image. Prior intrinsic-based pipelines consume \emph{all} intrinsic conditions \cite{luo2024intrinsic,zeng2024rgbx}, and resolving any conflicts that arise between them during editing is required for successful results, on a case-by-case basis \cite{lyu2025intrinsicedit}. Our pipeline does not suffer from such conflicts and provides a much more streamlined solution. We also show that the same formulation can be applied beyond intrinsic-space editing.

\section{Results}

We now present an evaluation of our method to demonstrate its capabilities. Our main applications are intrinsic-based editing and relighting, and we also show proof-of-concept text-based editing. The supplemental document includes additional results.

\Paragraph{Model}

We adopt Stable Diffusion 3.5 (SD3.5) Medium \cite{esser2024scaling}, which is a diffusion transformer with joint multimodal attention between text- and image-token streams. This attention mechanism provides general and powerful conditioning to the text stream, allowing us to use (a portion of) it to optimize a strong residual. We run the model with 30 inference steps.

We fine-tune the base SD3.5 model by augmenting it with intrinsic and lighting conditions. We add latent-encoded intrinsic conditions---albedo, normal, and roughness---by concatenating them to the noisy latent as additional channels, adjusting the input network layer accordingly \cite{zeng2024rgbx}. We repurpose the SD3.5 text-token stream to host the residual tokens, while the editable intrinsic condition remains in the spatial latent-input branch. For lighting, we use the recently proposed UniLight tokens \cite{zhang2025unilight}, a learned lighting representation that can be obtained from text or a reference image. We use pre-trained UniLight encoders kindly provided by the authors, which produce an embedding of 8 tokens of dimension 512. We expand those tokens to the 4096 token dimension of SD3.5 and append them to the 333-token text stream, with added learned positional embeddings.

Our training dataset comprises synthetic renders from Unreal and Evermotion scenes, together with a collection of real stock images, covering both indoor and outdoor environments. The intrinsic channels are estimated using an \rgbx model~\cite{zeng2024rgbx}; evaluation images are held out from training. Each image is additionally annotated with a UniLight embedding to encode its lighting. We apply intrinsic and UniLight dropout during training to teach the model to handle individual conditions at test time.

\Paragraph{Residual optimization}

For intrinsic-based editing and relighting, our residual embedding occupies all 333 tokens of the text stream. We initialize them with the embeddings of an empty string and run our residual optimization for 400-800 steps. The adversarial probe is an MLP that predicts the input condition from the residual tokens. We first pool the residual tokens to 16 tokens using adaptive 1D pooling, flatten the result, and feed it to two hidden layers of size 1024 with GeLU activations. Since the MLP aims to predict the input condition, its output layer is adjusted accordingly: to the size of the latent encoding of intrinsic conditions or to $8\times512$ for the UniLight condition. We balance the objectives in \cref{eq:minmax} with default $\lambda = 0.015$ (up to $0.05$ in some cases), and use AdamW optimizers with default parameters and learning rates of 5e-3 for the adversary probe and 0.1 for the residual. To stabilize training, we clip the gradient norms to 1.0. For relighting, the UniLight tokens are appended to the conditioning stream and kept fixed during optimization. We provide a qualitative sensitivity analysis for the adversarial weight in the supplemental document.

For text-based editing we use 40 residual tokens, appended to the original stream. Our proof-of-concept implementation supports modifying one word, so the probe predicts that word's token.

\begin{figure}[t]
    \centering
    \includegraphics[width=\linewidth]{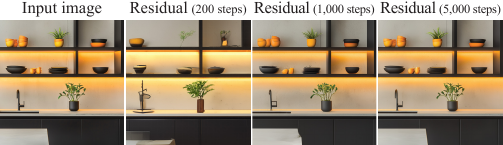}%
    \vspace{-1.5mm}
    \caption{
        \textbf{Residual capacity.} The strong conditioning power of the SD3.5 text-token stream enables our residual to accurately reconstruct the entire input image, when optimized long enough with only reconstruction loss and no additional model condition. We show images rendered with random noise, conditioned only by the residual, optimized for a different number of steps. This high-quality reconstruction ability translates to significantly reduced reliance on inversion when editing.
    }
    \vspace{-2mm}
    \label{fig:residual_capacity}
\end{figure}

We observe that SD3.5 has strong conditioning power, enabling a 333-token residual to memorize the input image with high accuracy when optimized for reconstruction only and without additional condition. We illustrate this in \cref{fig:residual_capacity}.

\begin{figure*}[t]
    \centering
    \includegraphics[width=\linewidth]{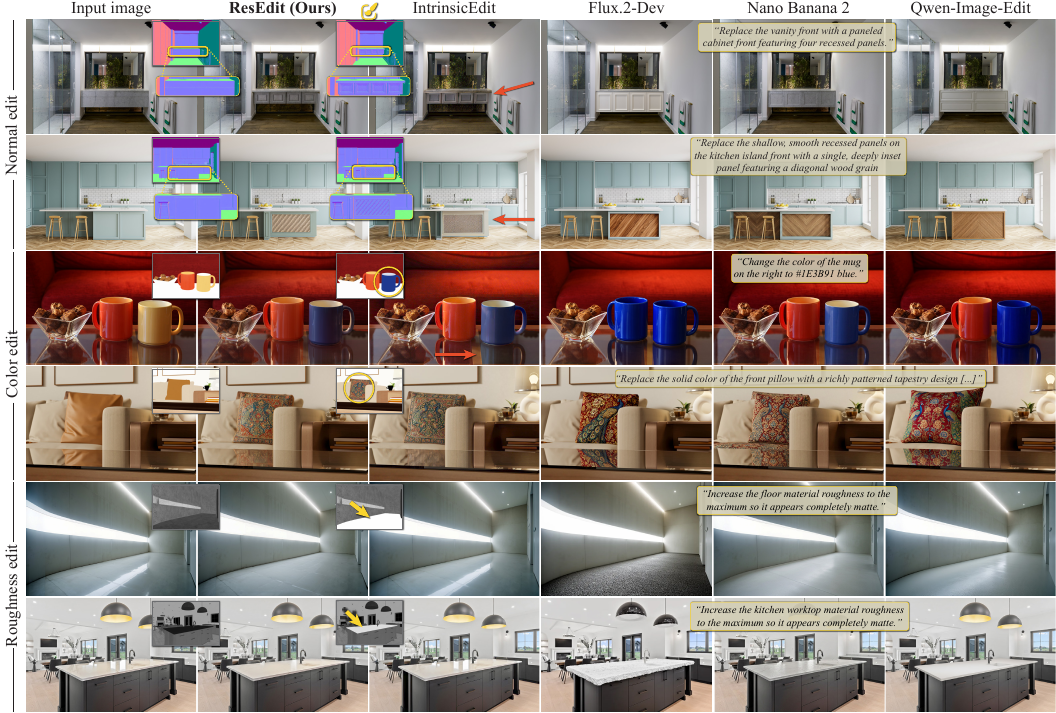}%
    \vspace{-1.5mm}
    \caption{
        \textbf{Material editing.} We showcase diverse surface appearance edits on intrinsic channels, including normal (top), albedo (middle), and roughness (bottom). Our method (2nd col.) produces more plausible results that better align with the user edit than IntrinsicEdit (3rd col.). Text-based methods (right) struggle to produce precise edits due to the limitations of natural language. 
        Best viewed zoomed-in.
    }
    \label{fig:material_editing}
    \vspace{-4mm}
\end{figure*}

\Paragraph{Intrinsic-space editing}

Our method consumes one external condition---input and edited intrinsic; we compare it against IntrinsicEdit~\cite{lyu2025intrinsicedit}, which requires all intrinsic channels. We also include several recent editing methods: Flux.2-Dev~\cite{flux-2-2025}, Nano Banana 2~\cite{google2026nanobanana2}, and Qwen-Image-Edit~\cite{wu2025qwenimagetechnicalreport}. We prompt these models via text for consistency, noting that some may additionally support visual references such as intrinsic maps. We provide expanded qualitative comparisons to Flux.1-Kontext \cite{FLUX1context} and Nano Banana \cite{google2025nanobanana} in the supplemental document.

In \cref{fig:material_editing}, we focus first on geometry editing via normal-map manipulation (top). Text-based methods deliver compelling results but natural language is too limited to describe precise geometric details required for precision editing. IntrinsicEdit offers closer adherence to user intent but struggles with content preservation outside the edit (1st row) and with accurately representing the specified geometry (2nd row). Our ResEdit better follows the specified edit while preserving unedited regions. Similar patterns emerge with albedo edits (rows 3-4): text-based methods can produce realistic-looking results but often struggle to faithfully reproduce the desired color or texture. IntrinsicEdit has difficulty with long-range effects such as reflections on the glossy table, whereas our method produces a more plausible result. We also show the ability to alter surface roughness, converting shiny surfaces into matte surfaces (rows 5-6). Even though this task is easier to describe in natural language, text-based methods are unable to faithfully make the material adjustment. IntrinsicEdit completely fails to change the surface roughness, indicating editing stiffness due to feature baking during noise inversion.

\begin{figure*}
    \centering
    \includegraphics[width=\linewidth]{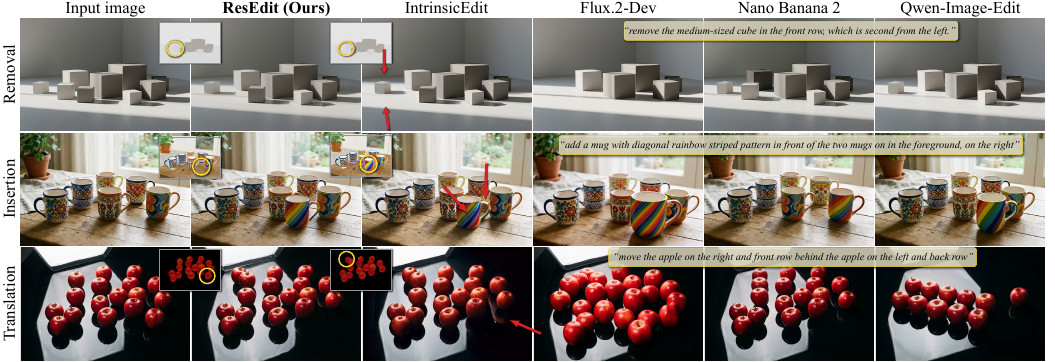}%
    \vspace{-1.5mm}
    \caption{
        \textbf{Editing examples}. We show edits for object removal, insertion, and translation (rows), respectively. The initial and edited intrinsic channels (e.g., edited albedo) are shown in the inset. Note that, compared to IntrinsicEdit, our method faithfully reconstructs unedited areas while resynthesizing plausible details in edited regions (e.g., shadows and reflections). Comparing to two text-based methods, Flux.1-Kontext and Nano Banana (rightmost columns), shows that text-based editing does not allow for precise editing (e.g., the rainbow pattern on the mug) and does not faithfully encode sufficiently detailed descriptions (apple example).
    }
    \label{fig:removal_insertion}
    \vspace{-6mm}
\end{figure*}

In \cref{fig:removal_insertion}, we assess our method's capacity for object manipulation, including removal (top), insertion (middle), and translation (bottom). As observed previously with surface editing, text-based methods struggle with edit precision. IntrinsicEdit successfully performs the desired edits, but compromises fidelity in unedited areas, evidenced by lost shadows (top, middle) or leftover reflections from a removed object (bottom). Conversely, our method faithfully applies the user's edit to the images.

\Paragraph{Relighting}

\begin{figure*}
    \centering
    \includegraphics[width=\linewidth]{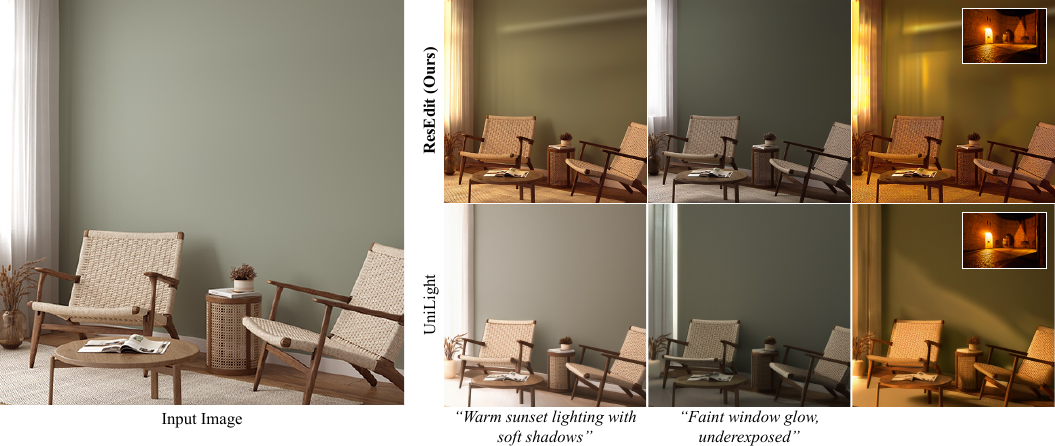}%
    \vspace{-2mm}
    \caption{
        \textbf{Relighting}. Given a lighting description---in the form of text prompt (first two examples) or reference illumination (rightmost example)---we encode it into the UniLight latent space \cite{zhang2025unilight}. We then use the resulting lighting tokens as input conditions for our method, achieving plausible, realistic relighting. The vanilla UniLight approach struggles with identity drifts (e.g., the vase material in the bottom left) as it is based on a simple \rgbxrgb pipeline which relies solely on intrinsic channels for identity preservation.
    }
    \label{fig:relighting}
    \vspace{-4mm}
\end{figure*}

We demonstrate our model's amenability to additional control signals via the UniLight embedding \cite{zhang2025unilight}. In \cref{fig:teaser,fig:relighting} we show reference-image-based relighting, confirming that both our method and UniLight alone successfully achieve the user's requested lighting transfer. In some cases, UniLight produces equally plausible or slightly better illumination. However, our method provides more consistent identity preservation (e.g., the carpet texture, the vase material on the left) and better texture retention (e.g., the plant on the left). UniLight's pipeline is based on a vanilla \rgbxrgb approach \cite{zeng2024rgbx} which generally struggles with identity preservation \cite{lyu2025intrinsicedit}.

\Paragraph{Feed-forward residual estimation}

\begin{figure}[t]
    \centering
    \includegraphics[width=\linewidth]{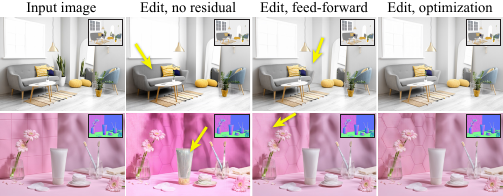}%
    \vspace{-1.5mm}
    \caption{
        \textbf{Feed-forward residual estimation.} Comparison between our residual inferred with a feed-forward encoder and a fully optimized residual on object removal (flower pot, top) and normal editing (wall structure, bottom).
    }
    \label{fig:residual_encoder}
\end{figure}

Optimizing our residual embedding can be time-consuming, preventing its use in a real-time context. To accelerate this step, we experimented with estimating our residual embedding directly from the input image using a feed-forward encoder based on a DINOv2-Large backbone \cite{oquab2024dinov2} and a transformer-based probe module (see supplemental document). These two networks are trained on our intrinsic-space dataset, using the same objectives as in \cref{eq:minmax} but propagating gradients through the residual into the encoder. We show results in \cref{fig:residual_encoder}. This approach adds negligible overhead over a vanilla inversion-only approach but provides substantial improvement, close to the full-optimization reference. Some slight degradation is visible, though the overall quality is maintained. The residual encoder runs in only 0.2 seconds (NVIDIA\,H100), compared to 2 minutes for our on-the-fly optimization. This approach provides a good trade-off, paving the way for fast editing.

\begin{table}[t]
    \centering
    \caption{
        For ground-truth image pairs, we compute the difference between optimized residuals. Adversarial optimization yields more similar residuals for before- and after-edit images, indicating better disentanglement.
    }
    \vspace{-0.5mm}%
    \begin{tabular}{@{}llll@{}}
        \toprule
        \textbf{Optimization}        & \textbf{MSE}\,$\downarrow$                    & \textbf{RMSE}\,$\downarrow$                   & \textbf{Cosine sim.}\,$\uparrow$            \\
        \midrule
        Without adversary~~~~~~~~~~~ & 1.9045        & 1.3517           & 0.2315           \\
        With adversary  & \textbf{1.7660} & \textbf{1.3191} & \textbf{0.2361  } \\
        \bottomrule
    \end{tabular}%
    \label{tab:adversary_metrics}
\end{table}

\begin{table}[t]
    \centering
    \vspace{-2mm}
    \caption{
        Quantitative metrics on albedo editing, roughness editing, and object removal against ground truths. Our full method achieves the best results. The FLIP$^*$ metrics are computed only on pixels whose intrinsics have not been edited.
    }
    \vspace{-0.5mm}
    \resizebox{\linewidth}{!}{%
        \setlength{\tabcolsep}{2pt}%
        \begin{tabular}{@{}llllll@{}}
            \toprule
            \textbf{Method}        & \textbf{PSNR (dB)}\,$\uparrow$                    & \textbf{SSIM}\,$\uparrow$ & \textbf{LPIPS}\,$\downarrow$ & \textbf{FLIP}\,$\downarrow$ & \textbf{FLIP$^*$}\,$\downarrow$                       \\
            \midrule
            IntrinsicEdit & 24.3016  & 0.8022   & 0.1069 & 0.1910 & 0.1588  \\
            \textbf{Ours} w/o adversary~~~~~~~~ & 24.9087            & 0.8258    & 0.0999 & 0.1557 & 0.1147     \\
            \textbf{Ours} feed-forward~~~~~~~~ & 23.5155            & 0.8121    & 0.1383 & 0.2049 & 0.1748   \\
            \textbf{Ours} w/ DDIM inv.~~~~~~~~ & 24.0385            & 0.8178    & 0.1043  & 0.2057 & 0.1694  \\
            \textbf{Ours} w/ 256 tokens~~~~~~~~ & 24.4316            & 0.8140    & 0.0997   & 0.1609 & 0.1190 \\
            \textbf{Ours} w/ 128 tokens~~~~~~~~ & 23.5391            & 0.8132    & 0.1023   & 0.2171 & 0.1809 \\
            \textbf{Ours} w/ 64 tokens~~~~~~~~ & 22.9612           & 0.8088    & 0.1164  & 0.2230 & 0.1924  \\
            \textbf{Ours} w/ weak probe~~~~~~~~ & 24.4556            & 0.8136    & 0.0997  & 0.1598 & 0.1183 \\
            \textbf{Ours} w/ strong probe~~~~~~~~ & 23.4706            & 0.8124    & 0.1024  & 0.2201 & 0.1871  \\
            \textbf{Ours} full  & \textbf{25.0554 } & \textbf{0.8260 } & \textbf{0.0992} & \textbf{0.1552} & \textbf{0.1128} \\
            \bottomrule
        \end{tabular}%
    }
    \label{tab:edit_metrics}
    \vspace{-3mm}
\end{table}

\Paragraph{Quantitative evaluation}

\begin{figure}[t]
    \centering
    \includegraphics[width=\linewidth]{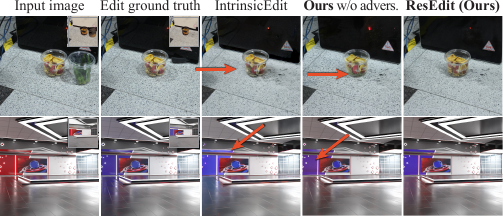}%
    \vspace{-1.5mm}
    \caption{
        \textbf{Adversary ablation.} Without adversarial optimization, our result improves over IntrinsicEdit but suffers from artifacts (top) and input-image leakage (bottom). With adversary, the better disentangled residual enables stronger, identity-preserving edits.
    }
    \label{fig:adversary_ablation}
\end{figure}

We quantitatively evaluate the disentanglement of the residual from the edited condition on a synthetic paired dataset including 10 albedo, 4 roughness, and 12 removal edits, exemplified in \cref{fig:adversary_ablation}. Specifically, we optimize two residuals using the before/after edit pairs and then measure the similarity between the optimized residuals. If the residuals are well disentangled from the conditioning signal, the residual difference should be small, since the observed image differences can be attributed primarily to changes in the condition. As shown in~\cref{tab:adversary_metrics}, adversarial optimization yields more disentangled residuals, achieving higher similarity (or lower difference) across all metrics.

We additionally evaluate edit fidelity on the same dataset using PSNR, SSIM, and LPIPS, and report FLIP~\cite{Andersson2020FLIP} to measure perceptual image differences. We also include FLIP metrics computed only on pixels whose intrinsics have not been edited. \Cref{tab:edit_metrics} compares against IntrinsicEdit and includes ablations over residual estimation, inversion strategy, residual token count, and adversarial probe capacity. Our full method achieves the best overall edit fidelity and perceptual error. We provide qualitative comparisons for the residual token-count and probe-capacity ablations, as well as FLIP error maps, in the supplemental document. 

\Paragraph{Intrinsic consistency analysis}

\begin{figure}
        \centering
        \includegraphics[width=0.47\textwidth]{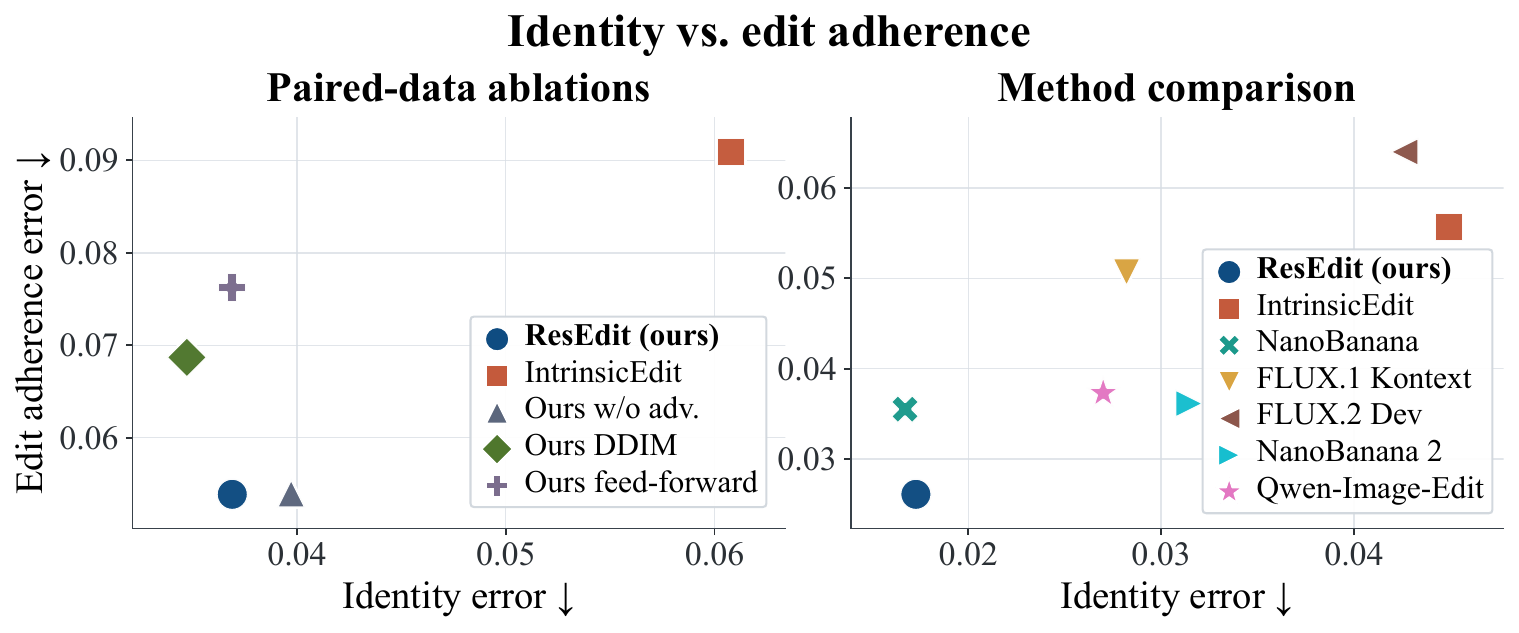}%
        \caption{
            \textbf{Intrinsic-consistency trade-off analysis.}
            The two axes show edit error and identity error, with lower values preferred. Our full method lies on the favorable trade-off frontier, achieving the lowest edit error and identity preservation comparable to the strongest baseline.
        }
        \label{fig:tradeoff_analysis}
\end{figure}

To further evaluate whether the generated RGB image faithfully realizes the intrinsic edit while preserving the remaining scene properties, we measure the intrinsic consistency using an \rgbx estimator. For the edited channel, we report the mean absolute error (MAE) between the target edited condition and the corresponding intrinsic estimate obtained from the output image. For the non-edited channels, we report the average MAE between their estimates before and after editing. The former measures edit adherence, while the latter measures preservation of intrinsic identity; lower values on both axes indicate a better trade-off. \Cref{tab:consistency_synthetic} summarizes this analysis, and \cref{fig:tradeoff_analysis} visualizes the corresponding trade-offs. Compared to all baselines and all variants, our ResEdit yields the lowest edit error while maintaining comparable identity error, achieving the most favorable trade-off.

\begin{table}[t]
    \centering
    \caption{
        Intrinsic consistency ablations on synthetic benchmarks (top rows) and comparisons on 20 synthetic/real edits (bottom rows). Our full method achieves the best edit accuracy while maintaining low identity error.
    }
    \vspace{-0.5mm}
        \setlength{\tabcolsep}{6pt} %
        \begin{tabular*}{\columnwidth}{@{}l@{\extracolsep{\fill}}ll@{}}
            \toprule
            \textbf{Method}        & \textbf{Edit error}\,$\downarrow$                    & \textbf{Identity error}\,$\downarrow$                            \\
            \midrule
            IntrinsicEdit & 0.0910        & 0.0608                   \\
            \textbf{Ours} feed-forward  & 0.0763        & 0.0369                   \\
            \textbf{Ours} w/ DDIM inversion  & 0.0687        & \textbf{0.0347}                   \\
            \textbf{Ours} w/o adversary  & 0.0540        & 0.0397                   \\
            \textbf{Ours} full  & \textbf{0.0539}        & 0.0369                   \\
            \midrule
            IntrinsicEdit & 0.0557        & 0.0449                   \\
            Nano Banana  & 0.0355        & \textbf{0.0167}                   \\
            Nano Banana 2  & 0.0361        & 0.0315                   \\
            Flux.1-Kontext  & 0.0506        & 0.0282                   \\
            Flux.2-Dev  & 0.0640        & 0.0426                   \\
            Qwen-Image-Edit  & 0.0373        & 0.0270                   \\
            \textbf{Ours} full  & \textbf{0.0261}        & 0.0173                   \\
            \bottomrule
        \end{tabular*}%
    \label{tab:consistency_synthetic}
\end{table}

\Paragraph{Text-based editing}

\begin{figure}[t]
    \centering
    \includegraphics[width=\linewidth]{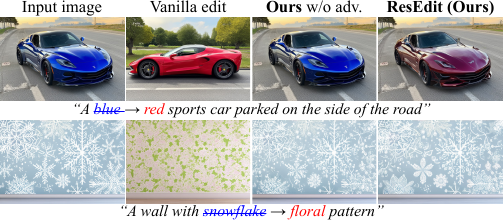}%
    \vspace{-1.5mm}
    \caption{
        \textbf{Text-based editing.}
        Simply inverting an input image and replacing the prompt leads to drastic content change (middle left). Our residual anchors the model to the original image (middle right), and adversarial optimization is crucial to avoid the leaking of edited attributes into the residual (right).
    }
    \label{fig:text_based_editing}
    \vspace{-4mm}
\end{figure}

As a preliminary extension beyond intrinsic-space editing, we show a proof-of-concept zero-shot text-based editing application on the vanilla, text-conditioned SD3.5 model. The input condition to our method is a text prompt, and the adversary probe in our residual optimization is trained to predict the word being changed, shown in blue in \cref{fig:text_based_editing}. A straightforward, ``vanilla edit'' approach involves inverting the input image to noise and re-synthesizing it with the edited prompt. This method yields content that is completely different from the input image, failing to achieve the goal of image editing. Our residual optimization anchors the model to the image and is able to change color or texture pattern while preserving input identity. Adversarial optimization is crucial to achieve a successful edit here; without it, the residual encodes the property that is being changed, e.g., ``blue'', and blocks editing by overpowering the edited prompt.


\Paragraph{Limitations}

ResEdit provides a streamlined intrinsic-space editing pipeline and improves the trade-off between edit adherence and identity preservation; however, it does not fully eliminate this trade-off. The residual embedding has high capacity, and although the adversarial objective reduces entanglement, some leakage can remain, especially with longer optimization or difficult edits. As a result, optimization length, adversarial weight, and probe capacity must be chosen to balance reconstruction fidelity against editability; the adversarial optimization can also be unstable. Furthermore, the per-image residual optimization is relatively slow (2--3 minutes on an NVIDIA\,H100 GPU). Our feed-forward encoder reduces this cost substantially, but currently achieves lower quality than full optimization. Our implementation inherits limitations from the underlying conditional diffusion backbone, including resolution, prompt-length constraints, and content biases, and depends on the quality of the intrinsic estimates used for training and evaluation. Finally, while our text-based editing examples suggest that the residual formulation can extend beyond intrinsic editing, these results are preliminary and less reliable than our intrinsic-space editing results.

\section{Conclusion}

Our ResEdit approach leverages generative imaging models for precise editing via intrinsic channels. Our key insight is to augment the generative model with a \emph{residual embedding} that encodes identity information about the image being edited. We optimize this residual to keep its content disentangled from the edited channel(s), thereby preserving identity and ensuring edit adherence. We accelerate the initialization of this residual with a feed-forward network, and demonstrate that the same approach can be extended to relighting and text-based editing. Our experimental results show state-of-the-art performance on intrinsic-space editing.


\bibliographystyle{eg-alpha-doi}
\bibliography{references}

@inproceedings{ho2020denoising,
  author       = {Jonathan Ho and
                  Ajay Jain and
                  Pieter Abbeel},
  editor       = {Hugo Larochelle and
                  Marc'Aurelio Ranzato and
                  Raia Hadsell and
                  Maria{-}Florina Balcan and
                  Hsuan{-}Tien Lin},
  title        = {Denoising Diffusion Probabilistic Models},
  booktitle    = {Advances in Neural Information Processing Systems 33: Annual Conference
                  on Neural Information Processing Systems 2020, NeurIPS 2020, December
                  6-12, 2020, virtual},
  year         = {2020},
  url          = {https://proceedings.neurips.cc/paper/2020/hash/4c5bcfec8584af0d967f1ab10179ca4b-Abstract.html},
  bibsource    = {dblp computer science bibliography, https://dblp.org}
}

@inproceedings{rombach2022high,
  author       = {Robin Rombach and
                  Andreas Blattmann and
                  Dominik Lorenz and
                  Patrick Esser and
                  Bj{\"{o}}rn Ommer},
  title        = {High-Resolution Image Synthesis with Latent Diffusion Models},
  booktitle    = {{IEEE/CVF} Conference on Computer Vision and Pattern Recognition,
                  {CVPR} 2022, New Orleans, LA, USA, June 18-24, 2022},
  pages        = {10674--10685},
  publisher    = {{IEEE}},
  year         = {2022},
  url          = {https://doi.org/10.1109/CVPR52688.2022.01042},
  doi          = {10.1109/CVPR52688.2022.01042},
  bibsource    = {dblp computer science bibliography, https://dblp.org}
}

@inproceedings{esser2024scaling,
  author       = {Patrick Esser and
                  Sumith Kulal and
                  Andreas Blattmann and
                  Rahim Entezari and
                  Jonas M{\"{u}}ller and
                  Harry Saini and
                  Yam Levi and
                  Dominik Lorenz and
                  Axel Sauer and
                  Frederic Boesel and
                  Dustin Podell and
                  Tim Dockhorn and
                  Zion English and
                  Robin Rombach},
  title        = {Scaling Rectified Flow Transformers for High-Resolution Image Synthesis},
  booktitle    = {Forty-first International Conference on Machine Learning, {ICML} 2024,
                  Vienna, Austria, July 21-27, 2024},
  publisher    = {OpenReview.net},
  year         = {2024},
  url          = {https://openreview.net/forum?id=FPnUhsQJ5B},
  bibsource    = {dblp computer science bibliography, https://dblp.org}
}

@inproceedings{meng2022sdedit,
  author       = {Chenlin Meng and
                  Yutong He and
                  Yang Song and
                  Jiaming Song and
                  Jiajun Wu and
                  Jun{-}Yan Zhu and
                  Stefano Ermon},
  title        = {SDEdit: Guided Image Synthesis and Editing with Stochastic Differential
                  Equations},
  booktitle    = {The Tenth International Conference on Learning Representations, {ICLR}
                  2022, Virtual Event, April 25-29, 2022},
  publisher    = {OpenReview.net},
  year         = {2022},
  url          = {https://openreview.net/forum?id=aBsCjcPu\_tE},
  bibsource    = {dblp computer science bibliography, https://dblp.org}
}

@inproceedings{hertz2023prompt,
  author       = {Amir Hertz and
                  Ron Mokady and
                  Jay Tenenbaum and
                  Kfir Aberman and
                  Yael Pritch and
                  Daniel Cohen{-}Or},
  title        = {Prompt-to-Prompt Image Editing with Cross-Attention Control},
  booktitle    = {The Eleventh International Conference on Learning Representations,
                  {ICLR} 2023, Kigali, Rwanda, May 1-5, 2023},
  publisher    = {OpenReview.net},
  year         = {2023},
  url          = {https://openreview.net/forum?id=\_CDixzkzeyb},
  bibsource    = {dblp computer science bibliography, https://dblp.org}
}

@inproceedings{cao2023masactrl,
  author       = {Mingdeng Cao and
                  Xintao Wang and
                  Zhongang Qi and
                  Ying Shan and
                  Xiaohu Qie and
                  Yinqiang Zheng},
  title        = {MasaCtrl: Tuning-Free Mutual Self-Attention Control for Consistent
                  Image Synthesis and Editing},
  booktitle    = {{IEEE/CVF} International Conference on Computer Vision, {ICCV} 2023,
                  Paris, France, October 1-6, 2023},
  pages        = {22503--22513},
  publisher    = {{IEEE}},
  year         = {2023},
  url          = {https://doi.org/10.1109/ICCV51070.2023.02062},
  doi          = {10.1109/ICCV51070.2023.02062},
  bibsource    = {dblp computer science bibliography, https://dblp.org}
}

@inproceedings{zhang2023adding,
  author       = {Lvmin Zhang and
                  Anyi Rao and
                  Maneesh Agrawala},
  title        = {Adding Conditional Control to Text-to-Image Diffusion Models},
  booktitle    = {{IEEE/CVF} International Conference on Computer Vision, {ICCV} 2023,
                  Paris, France, October 1-6, 2023},
  pages        = {3813--3824},
  publisher    = {{IEEE}},
  year         = {2023},
  url          = {https://doi.org/10.1109/ICCV51070.2023.00355},
  doi          = {10.1109/ICCV51070.2023.00355},
  bibsource    = {dblp computer science bibliography, https://dblp.org}
}

@inproceedings{zhu2025training,
  title={Training-Free Diffusion for Geometric Image Editing}, 
  author={Zhu, Hanshen and Zhu, Zhen and Zhang, Kaile and Gong, Yiming and Liu, Yuliang and Bai, Xiang},
  booktitle={ICCV}, 
  year={2025}
}

@article{FLUX1context,
  author       = {{Black Forest Labs}},
  title        = {{FLUX.1} Kontext: Flow Matching for In-Context Image Generation and
                  Editing in Latent Space},
  journal      = {CoRR},
  volume       = {abs/2506.15742},
  year         = {2025},
  url          = {https://doi.org/10.48550/arXiv.2506.15742},
  doi          = {10.48550/ARXIV.2506.15742},
  eprinttype    = {arXiv},
  eprint       = {2506.15742},
  bibsource    = {dblp computer science bibliography, https://dblp.org}
}

@inproceedings{song2021denoising,
  author       = {Jiaming Song and
                  Chenlin Meng and
                  Stefano Ermon},
  title        = {Denoising Diffusion Implicit Models},
  booktitle    = {9th International Conference on Learning Representations, {ICLR} 2021,
                  Virtual Event, Austria, May 3-7, 2021},
  publisher    = {OpenReview.net},
  year         = {2021},
  url          = {https://openreview.net/forum?id=St1giarCHLP},
  bibsource    = {dblp computer science bibliography, https://dblp.org}
}

@inproceedings{hong2024on,
  author       = {Seongmin Hong and
                  Kyeonghyun Lee and
                  Suh Yoon Jeon and
                  Hyewon Bae and
                  Se Young Chun},
  title        = {On Exact Inversion of DPM-Solvers},
  booktitle    = {{IEEE/CVF} Conference on Computer Vision and Pattern Recognition,
                  {CVPR} 2024, Seattle, WA, USA, June 16-22, 2024},
  pages        = {7069--7078},
  publisher    = {{IEEE}},
  year         = {2024},
  url          = {https://doi.org/10.1109/CVPR52733.2024.00675},
  doi          = {10.1109/CVPR52733.2024.00675},
  bibsource    = {dblp computer science bibliography, https://dblp.org}
}

@inproceedings{mokady2023null,
  author       = {Ron Mokady and
                  Amir Hertz and
                  Kfir Aberman and
                  Yael Pritch and
                  Daniel Cohen{-}Or},
  title        = {Null-text Inversion for Editing Real Images using Guided Diffusion
                  Models},
  booktitle    = {{IEEE/CVF} Conference on Computer Vision and Pattern Recognition,
                  {CVPR} 2023, Vancouver, BC, Canada, June 17-24, 2023},
  pages        = {6038--6047},
  publisher    = {{IEEE}},
  year         = {2023},
  url          = {https://doi.org/10.1109/CVPR52729.2023.00585},
  doi          = {10.1109/CVPR52729.2023.00585},
  bibsource    = {dblp computer science bibliography, https://dblp.org}
}

@inproceedings{miyake2025negative,
  author       = {Daiki Miyake and
                  Akihiro Iohara and
                  Yu Saito and
                  Toshiyuki Tanaka},
  title        = {Negative-Prompt Inversion: Fast Image Inversion for Editing with Text-Guided
                  Diffusion Models},
  booktitle    = {{IEEE/CVF} Winter Conference on Applications of Computer Vision, {WACV}
                  2025, Tucson, AZ, USA, February 26 - March 6, 2025},
  pages        = {2063--2072},
  publisher    = {{IEEE}},
  year         = {2025},
  url          = {https://doi.org/10.1109/WACV61041.2025.00207},
  doi          = {10.1109/WACV61041.2025.00207},
  bibsource    = {dblp computer science bibliography, https://dblp.org}
}

@inproceedings{dong2023prompt,
  author       = {Wenkai Dong and
                  Song Xue and
                  Xiaoyue Duan and
                  Shumin Han},
  title        = {Prompt Tuning Inversion for Text-Driven Image Editing Using Diffusion
                  Models},
  booktitle    = {{IEEE/CVF} International Conference on Computer Vision, {ICCV} 2023,
                  Paris, France, October 1-6, 2023},
  pages        = {7396--7406},
  publisher    = {{IEEE}},
  year         = {2023},
  url          = {https://doi.org/10.1109/ICCV51070.2023.00683},
  doi          = {10.1109/ICCV51070.2023.00683},
  bibsource    = {dblp computer science bibliography, https://dblp.org}
}

@inproceedings{chung2024prompt,
  author       = {Hyungjin Chung and
                  Jong Chul Ye and
                  Peyman Milanfar and
                  Mauricio Delbracio},
  title        = {Prompt-tuning Latent Diffusion Models for Inverse Problems},
  booktitle    = {Forty-first International Conference on Machine Learning, {ICML} 2024,
                  Vienna, Austria, July 21-27, 2024},
  publisher    = {OpenReview.net},
  year         = {2024},
  url          = {https://openreview.net/forum?id=hrwIndai8e},
  bibsource    = {dblp computer science bibliography, https://dblp.org}
}

@inproceedings{
    na2025diffusion,
    title={Diffusion Adaptive Text Embedding for Text-to-Image Diffusion Models},
    author={Byeonghu Na and Minsang Park and Gyuwon Sim and Donghyeok Shin and HeeSun Bae and Mina Kang and Se Jung Kwon and Wanmo Kang and Il-chul Moon},
    booktitle={The Thirty-ninth Annual Conference on Neural Information Processing Systems},
    year={2025},
    url={https://openreview.net/forum?id=cHi8QxGrZH}
}

@inproceedings{nguyen2025swiftedit,
  author       = {Trong{-}Tung Nguyen and
                  Quang Nguyen and
                  Khoi Nguyen and
                  Anh Tuan Tran and
                  Cuong Pham},
  title        = {SwiftEdit: Lightning Fast Text-Guided Image Editing via One-Step Diffusion},
  booktitle    = {{IEEE/CVF} Conference on Computer Vision and Pattern Recognition,
                  {CVPR} 2025, Nashville, TN, USA, June 11-15, 2025},
  pages        = {21492--21501},
  publisher    = {Computer Vision Foundation / {IEEE}},
  year         = {2025},
  url          = {https://openaccess.thecvf.com/content/CVPR2025/html/Nguyen\_SwiftEdit\_Lightning\_Fast\_Text-Guided\_Image\_Editing\_via\_One-Step\_Diffusion\_CVPR\_2025\_paper.html},
  doi          = {10.1109/CVPR52734.2025.02002},
  bibsource    = {dblp computer science bibliography, https://dblp.org}
}

@inproceedings{chen2016infogan,
  author       = {Xi Chen and
                  Yan Duan and
                  Rein Houthooft and
                  John Schulman and
                  Ilya Sutskever and
                  Pieter Abbeel},
  editor       = {Daniel D. Lee and
                  Masashi Sugiyama and
                  Ulrike von Luxburg and
                  Isabelle Guyon and
                  Roman Garnett},
  title        = {InfoGAN: Interpretable Representation Learning by Information Maximizing
                  Generative Adversarial Nets},
  booktitle    = {Advances in Neural Information Processing Systems 29: Annual Conference
                  on Neural Information Processing Systems 2016, December 5-10, 2016,
                  Barcelona, Spain},
  pages        = {2172--2180},
  year         = {2016},
  url          = {https://proceedings.neurips.cc/paper/2016/hash/7c9d0b1f96aebd7b5eca8c3edaa19ebb-Abstract.html},
  bibsource    = {dblp computer science bibliography, https://dblp.org}
}

@incollection{ganin2017domain,
  author       = {Yaroslav Ganin and
                  Evgeniya Ustinova and
                  Hana Ajakan and
                  Pascal Germain and
                  Hugo Larochelle and
                  Fran{\c{c}}ois Laviolette and
                  Mario Marchand and
                  Victor S. Lempitsky},
  editor       = {Gabriela Csurka},
  title        = {Domain-Adversarial Training of Neural Networks},
  booktitle    = {Domain Adaptation in Computer Vision Applications},
  series       = {Advances in Computer Vision and Pattern Recognition},
  pages        = {189--209},
  publisher    = {Springer},
  year         = {2017},
  url          = {https://doi.org/10.1007/978-3-319-58347-1\_10},
  doi          = {10.1007/978-3-319-58347-1\_10},
  bibsource    = {dblp computer science bibliography, https://dblp.org}
}

@inproceedings{edwards2016censoring,
  author       = {Harrison Edwards and
                  Amos J. Storkey},
  editor       = {Yoshua Bengio and
                  Yann LeCun},
  title        = {Censoring Representations with an Adversary},
  booktitle    = {4th International Conference on Learning Representations, {ICLR} 2016,
                  San Juan, Puerto Rico, May 2-4, 2016, Conference Track Proceedings},
  year         = {2016},
  url          = {http://arxiv.org/abs/1511.05897},
  bibsource    = {dblp computer science bibliography, https://dblp.org}
}

@inproceedings{lample2017fader,
  author       = {Guillaume Lample and
                  Neil Zeghidour and
                  Nicolas Usunier and
                  Antoine Bordes and
                  Ludovic Denoyer and
                  Marc'Aurelio Ranzato},
  editor       = {Isabelle Guyon and
                  Ulrike von Luxburg and
                  Samy Bengio and
                  Hanna M. Wallach and
                  Rob Fergus and
                  S. V. N. Vishwanathan and
                  Roman Garnett},
  title        = {Fader Networks: Manipulating Images by Sliding Attributes},
  booktitle    = {Advances in Neural Information Processing Systems 30: Annual Conference
                  on Neural Information Processing Systems 2017, December 4-9, 2017,
                  Long Beach, CA, {USA}},
  pages        = {5967--5976},
  year         = {2017},
  url          = {https://proceedings.neurips.cc/paper/2017/hash/3fd60983292458bf7dee75f12d5e9e05-Abstract.html},
  bibsource    = {dblp computer science bibliography, https://dblp.org}
}

@inproceedings{wu2023uncovering,
  author       = {Qiucheng Wu and
                  Yujian Liu and
                  Handong Zhao and
                  Ajinkya Kale and
                  Trung Bui and
                  Tong Yu and
                  Zhe Lin and
                  Yang Zhang and
                  Shiyu Chang},
  title        = {Uncovering the Disentanglement Capability in Text-to-Image Diffusion
                  Models},
  booktitle    = {{IEEE/CVF} Conference on Computer Vision and Pattern Recognition,
                  {CVPR} 2023, Vancouver, BC, Canada, June 17-24, 2023},
  pages        = {1900--1910},
  publisher    = {{IEEE}},
  year         = {2023},
  url          = {https://doi.org/10.1109/CVPR52729.2023.00189},
  doi          = {10.1109/CVPR52729.2023.00189},
  bibsource    = {dblp computer science bibliography, https://dblp.org}
}

@inproceedings{kocsis2024intrinsic,
  author       = {Peter Kocsis and
                  Vincent Sitzmann and
                  Matthias Nie{\ss}ner},
  title        = {Intrinsic Image Diffusion for Indoor Single-view Material Estimation},
  booktitle    = {{IEEE/CVF} Conference on Computer Vision and Pattern Recognition,
                  {CVPR} 2024, Seattle, WA, USA, June 16-22, 2024},
  pages        = {5198--5208},
  publisher    = {{IEEE}},
  year         = {2024},
  url          = {https://doi.org/10.1109/CVPR52733.2024.00497},
  doi          = {10.1109/CVPR52733.2024.00497},
  bibsource    = {dblp computer science bibliography, https://dblp.org}
}

@inproceedings{luo2024intrinsic,
  author       = {Jundan Luo and
                  Duygu Ceylan and
                  Jae Shin Yoon and
                  Nanxuan Zhao and
                  Julien Philip and
                  Anna Fr{\"{u}}hst{\"{u}}ck and
                  Wenbin Li and
                  Christian Richardt and
                  Tuanfeng Y. Wang},
  editor       = {Andres Burbano and
                  Denis Zorin and
                  Wojciech Jarosz},
  title        = {IntrinsicDiffusion: Joint Intrinsic Layers from Latent Diffusion Models},
  booktitle    = {{ACM} {SIGGRAPH} 2024 Conference Papers, {SIGGRAPH} 2024, Denver,
                  CO, USA, 27 July 2024- 1 August 2024},
  pages        = {74},
  publisher    = {{ACM}},
  year         = {2024},
  url          = {https://doi.org/10.1145/3641519.3657472},
  doi          = {10.1145/3641519.3657472},
  bibsource    = {dblp computer science bibliography, https://dblp.org}
}

@article{careaga2024colorful,
  author       = {Chris Careaga and
                  Yagiz Aksoy},
  title        = {Colorful Diffuse Intrinsic Image Decomposition in the Wild},
  journal      = {{ACM} Trans. Graph.},
  volume       = {43},
  number       = {6},
  pages        = {178:1--178:12},
  year         = {2024},
  url          = {https://doi.org/10.1145/3687984},
  doi          = {10.1145/3687984},
  bibsource    = {dblp computer science bibliography, https://dblp.org}
}

@inproceedings{zeng2024rgbx,
  author       = {Zheng Zeng and
                  Valentin Deschaintre and
                  Iliyan Georgiev and
                  Yannick Hold{-}Geoffroy and
                  Yiwei Hu and
                  Fujun Luan and
                  Ling{-}Qi Yan and
                  Milos Hasan},
  editor       = {Andres Burbano and
                  Denis Zorin and
                  Wojciech Jarosz},
  title        = {RGB{\(\leftrightarrow\)}X: Image decomposition and synthesis using
                  material- and lighting-aware diffusion models},
  booktitle    = {{ACM} {SIGGRAPH} 2024 Conference Papers, {SIGGRAPH} 2024, Denver,
                  CO, USA, 27 July 2024- 1 August 2024},
  pages        = {75},
  publisher    = {{ACM}},
  year         = {2024},
  url          = {https://doi.org/10.1145/3641519.3657445},
  doi          = {10.1145/3641519.3657445},
  bibsource    = {dblp computer science bibliography, https://dblp.org}
}

@article{lyu2025intrinsicedit,
  author       = {Linjie Lyu and
                  Valentin Deschaintre and
                  Yannick Hold{-}Geoffroy and
                  Milos Hasan and
                  Jae Shin Yoon and
                  Thomas Leimk{\"{u}}hler and
                  Christian Theobalt and
                  Iliyan Georgiev},
  title        = {IntrinsicEdit: Precise generative image manipulation in intrinsic
                  space},
  journal      = {{ACM} Trans. Graph.},
  volume       = {44},
  number       = {4},
  pages        = {106:1--106:13},
  year         = {2025},
  url          = {https://doi.org/10.1145/3731173},
  doi          = {10.1145/3731173},
  bibsource    = {dblp computer science bibliography, https://dblp.org}
}

@inproceedings{kocsis2024lightit,
  author       = {Peter Kocsis and
                  Julien Philip and
                  Kalyan Sunkavalli and
                  Matthias Nie{\ss}ner and
                  Yannick Hold{-}Geoffroy},
  title        = {LightIt: Illumination Modeling and Control for Diffusion Models},
  booktitle    = {{IEEE/CVF} Conference on Computer Vision and Pattern Recognition,
                  {CVPR} 2024, Seattle, WA, USA, June 16-22, 2024},
  pages        = {9359--9369},
  publisher    = {{IEEE}},
  year         = {2024},
  url          = {https://doi.org/10.1109/CVPR52733.2024.00894},
  doi          = {10.1109/CVPR52733.2024.00894},
  bibsource    = {dblp computer science bibliography, https://dblp.org}
}

@inproceedings{zhang2025zerocomp,
  author       = {Zitian Zhang and
                  Fr{\'{e}}d{\'{e}}ric Fortier{-}Chouinard and
                  Mathieu Garon and
                  Anand Bhattad and
                  Jean{-}Fran{\c{c}}ois Lalonde},
  title        = {Zerocomp: Zero-Shot Object Compositing from Image Intrinsics via Diffusion},
  booktitle    = {{IEEE/CVF} Winter Conference on Applications of Computer Vision, {WACV}
                  2025, Tucson, AZ, USA, February 26 - March 6, 2025},
  pages        = {483--494},
  publisher    = {{IEEE}},
  year         = {2025},
  url          = {https://doi.org/10.1109/WACV61041.2025.00057},
  doi          = {10.1109/WACV61041.2025.00057},
  bibsource    = {dblp computer science bibliography, https://dblp.org}
}

@inproceedings{brooks2023instructpix2pix,
  title={Instructpix2pix: Learning to follow image editing instructions},
  author={Brooks, Tim and Holynski, Aleksander and Efros, Alexei A},
  booktitle={Proceedings of the IEEE/CVF Conference on Computer Vision and Pattern Recognition},
  pages={18392--18402},
  year={2023}
}

@article{barrow1978recovering,
  title={Recovering intrinsic scene characteristics},
  author={Barrow, Harry and Tenenbaum, J and Hanson, A and Riseman, E},
  journal={Comput. vis. syst},
  volume={2},
  number={3-26},
  pages={2},
  year={1978}
}

@inproceedings{mahajan2024prompting,
  title={Prompting Hard or Hardly Prompting: Prompt Inversion for Text-to-Image Diffusion Models},
  author={Mahajan, Shweta and Rahman, Tanzila and Yi, Kwang Moo and Sigal, Leonid},
  booktitle={Proceedings of the IEEE/CVF Conference on Computer Vision and Pattern Recognition},
  pages={6808--6817},
  year={2024}
}

@article{ho2022classifier,
  title={Classifier-free diffusion guidance},
  author={Ho, Jonathan and Salimans, Tim},
  journal={arXiv preprint arXiv:2207.12598},
  year={2022}
}

@article{gal2022image,
  title={An image is worth one word: Personalizing text-to-image generation using textual inversion},
  author={Gal, Rinon and Alaluf, Yuval and Atzmon, Yuval and Patashnik, Or and Bermano, Amit H and Chechik, Gal and Cohen-Or, Daniel},
  journal={arXiv preprint arXiv:2208.01618},
  year={2022}
}

@inproceedings{ju2024pnp,
title={PnP Inversion: Boosting Diffusion-based Editing with 3 Lines of Code},
author={Xuan Ju and Ailing Zeng and Yuxuan Bian and Shaoteng Liu and Qiang Xu},
booktitle={The Twelfth International Conference on Learning Representations},
year={2024},
url={https://openreview.net/forum?id=FoMZ4ljhVw}
}

@misc{google2025nanobanana,
  author = {{Google DeepMind}},
  title = {Nano Banana: A Multimodal Diffusion Model for High-Fidelity Image Generation and Iterative Editing},
  year = {2025},
  url = {https://deepmind.google/technologies/gemini/},
  note = {Part of the Gemini model family}
}

@misc{liu2025step1xedit,
      title={Step1X-Edit: A Practical Framework for General Image Editing}, 
      author={Shiyu Liu and Yucheng Han and Peng Xing and Fukun Yin and Rui Wang and Wei Cheng and Jiaqi Liao and Yingming Wang and Honghao Fu and Chunrui Han and Guopeng Li and Yuang Peng and Quan Sun and Jingwei Wu and Yan Cai and Zheng Ge and Ranchen Ming and Lei Xia and Xianfang Zeng and Yibo Zhu and Binxing Jiao and Xiangyu Zhang and Gang Yu and Daxin Jiang},
      year={2025},
      eprint={2504.17761},
      archivePrefix={arXiv},
      primaryClass={cs.CV},
      url={https://arxiv.org/abs/2504.17761}, 
}

@misc{gpt4oimage,
  title={Addendum to the GPT-4o System Card: Native Image Generation and Editing},
  author={OpenAI},
  year={2025},
  howpublished={\url{https://openai.com/index/introducing-4o-image-generation/}},
  note={Accessed: 2026-01-16}
}

@inproceedings{dhariwal2021diffusion,
  title={Diffusion Models Beat GANs on Image Synthesis},
  author={Dhariwal, Prafulla and Nichol, Alexander},
  booktitle={Advances in Neural Information Processing Systems (NeurIPS)},
  volume={34},
  pages={8780--8794},
  year={2021}
}

@misc{wang2025seededit,
    title={SeedEdit 3.0: Fast and High-Quality Generative Image Editing}, 
    author={Peng Wang and Yichun Shi and Xiaochen Lian and Zhonghua Zhai and Xin Xia and Xuefeng Xiao and Weilin Huang and Jianchao Yang},
    year={2025},
    eprint={2506.05083},
    archivePrefix={arXiv},
    primaryClass={cs.CV},
    url={https://arxiv.org/abs/2506.05083}, 
}

@inproceedings{tumanyan2023plug,
  title={Plug-and-play diffusion features for text-driven image-to-image translation},
  author={Tumanyan, Narek and Geyer, Michal and Bagon, Shai and Dekel, Tali},
  booktitle={Proceedings of the IEEE/CVF Conference on Computer Vision and Pattern Recognition (CVPR)},
  pages={1921--1930},
  year={2023}
}

@inproceedings{ganin2015gradient,
  title     = {Unsupervised Domain Adaptation by Backpropagation},
  author    = {Ganin, Yaroslav and Lempitsky, Victor},
  booktitle = {Proceedings of the 32nd International Conference on Machine Learning},
  pages     = {1180--1189},
  year      = {2015},
  editor    = {Bach, Francis and Blei, David},
  volume    = {37},
  series    = {Proceedings of Machine Learning Research},
  address   = {Lille, France},
  month     = {07--09 Jul},
  publisher = {PMLR},
  url       = {https://proceedings.mlr.press/v37/ganin15.html}
}

@misc{zhang2025unilight,
    title={UniLight: A Unified Representation for Lighting},
    author={Zitian Zhang and Iliyan Georgiev and Michael Fischer and Yannick Hold-Geoffroy and Jean-François Lalonde and Valentin Deschaintre},
    year={2025},
    eprint={2512.04267},
    archivePrefix={arXiv},
    primaryClass={cs.CV},
    url={https://arxiv.org/abs/2512.04267},
}

@inproceedings{kulikov2025flowedit,
    title={Flowedit: Inversion-free text-based editing using pre-trained flow models},
    author={Kulikov, Vladimir and Kleiner, Matan and Huberman-Spiegelglas, Inbar and Michaeli, Tomer},
    booktitle={Proceedings of the IEEE/CVF International Conference on Computer Vision},
    pages={19721--19730},
    year={2025}
}

@inproceedings{samuel2024lightning,
  author    = {Dvir Samuel and Barak Meiri and Haggai Maron and Yoad Tewel and Nir Darshan and Shai Avidan and Gal Chechik and Rami Ben-Ari},
  title     = {Lightning-fast Image Inversion and Editing for Text-to-Image Diffusion Models},
  booktitle = {Proceedings of the International Conference on Learning Representations (ICLR)},
  year      = {2025}
}

@inproceedings{Lipman2023FlowMatching,
  title     = {Flow Matching for Generative Modeling},
  author    = {Lipman, Yaron and Chen, Ricky T. Q. and Ben-Hamu, Heli and Nickel, Maximilian and Le, Matt},
  booktitle = {International Conference on Learning Representations (ICLR)},
  year      = {2023},
  url       = {https://arxiv.org/abs/2210.02747}
}

@inproceedings{Liu2023RectifiedFlow,
  title     = {Flow Straight and Fast: Learning to Generate and Transfer Data with Rectified Flow},
  author    = {Liu, Xingchao and Gong, Lemeng and Liu, Qiang},
  booktitle = {International Conference on Learning Representations (ICLR)},
  year      = {2023},
  url       = {https://arxiv.org/abs/2209.03003}
}

@inproceedings{li2024source,
  title={Source Prompt Disentangled Inversion for Boosting Image Editability with Diffusion Models},
  author={Li, Ruibin and Li, Ruihuang and Guo, Song and Zhang, Lei},
  booktitle={European Conference on Computer Vision},
  year={2024}
}

@misc{ouyang2025proedit,
  title={ProEdit: Inversion-based Editing From Prompts Done Right},
  author={Ouyang, Zhi and Zheng, Dian and Wu, Xiao-Ming and Jiang, Jian-Jian and Lin, Kun-Yu and Meng, Jingke and Zheng, Wei-Shi},
  year={2025},
  eprint={2512.22118},
  archivePrefix={arXiv},
  primaryClass={cs.CV},
  url={https://arxiv.org/abs/2512.22118}
}

@article{hou2024high,
  title={High-Fidelity Diffusion-based Image Editing},
  author={Hou, Chen and Wei, Guoqiang and Chen, Zhibo},
  journal={arXiv preprint arXiv:2312.15707},
  year={2023}
}

@inproceedings{kim2018disentangling,
  title     = {Disentangling by Factorising},
  author    = {Kim, Hyunjik and Mnih, Andriy},
  booktitle = {Proceedings of the 35th International Conference on Machine Learning (ICML)},
  pages     = {2649--2658},
  year      = {2018},
  editor    = {Dy, Jennifer and Krause, Andreas},
  volume    = {80},
  series    = {Proceedings of Machine Learning Research},
  publisher = {PMLR},
  url       = {https://proceedings.mlr.press/v80/kim18b.html}
}

@misc{oquab2024dinov2,
      title={DINOv2: Learning Robust Visual Features without Supervision}, 
      author={Maxime Oquab and Timothée Darcet and Théo Moutakanni and Huy Vo and Marc Szafraniec and Vasil Khalidov and Pierre Fernandez and Daniel Haziza and Francisco Massa and Alaaeldin El-Nouby and Mahmoud Assran and Nicolas Ballas and Wojciech Galuba and Russell Howes and Po-Yao Huang and Shang-Wen Li and Ishan Misra and Michael Rabbat and Vasu Sharma and Gabriel Synnaeve and Hu Xu and Hervé Jegou and Julien Mairal and Patrick Labatut and Armand Joulin and Piotr Bojanowski},
      year={2024},
      eprint={2304.07193},
      archivePrefix={arXiv},
      primaryClass={cs.CV},
      url={https://arxiv.org/abs/2304.07193}, 
}

@InProceedings{alaluf2021restyle,
      author = {Alaluf, Yuval and Patashnik, Or and Cohen-Or, Daniel},
      title = {ReStyle: A Residual-Based StyleGAN Encoder via Iterative Refinement}, 
      month = {October},
      booktitle = {Proceedings of the IEEE/CVF International Conference on Computer Vision (ICCV)},  
      year = {2021}
}

@article{xu2026invert,
author = {Xu, Yangyang and Shao, Wenqi and Du, Yong and Zhu, Haiming and Zhou, Yang and Xie, Jiayuan and Luo, Ping and He, Shengfeng},
title = {Invert Your Prompt: Editing-Aware Diffusion Inversion},
year = {2026},
issue_date = {Mar 2026},
publisher = {Kluwer Academic Publishers},
address = {USA},
volume = {134},
number = {4},
issn = {0920-5691},
doi = {10.1007/s11263-025-02691-1},
journal = {Int. J. Comput. Vision},
month = mar,
numpages = {18}
}

@misc{flux-2-2025,
    author={{Black Forest Labs}},
    title={{FLUX.2: Frontier Visual Intelligence}},
    year={2025},
    howpublished={\url{https://bfl.ai/blog/flux-2}},
}

@misc{wu2025qwenimagetechnicalreport,
      title={Qwen-Image Technical Report}, 
      author={Chenfei Wu and Jiahao Li and Jingren Zhou and Junyang Lin and Kaiyuan Gao and Kun Yan and Sheng-ming Yin and Shuai Bai and Xiao Xu and Yilei Chen and Yuxiang Chen and Zecheng Tang and Zekai Zhang and Zhengyi Wang and An Yang and Bowen Yu and Chen Cheng and Dayiheng Liu and Deqing Li and Hang Zhang and Hao Meng and Hu Wei and Jingyuan Ni and Kai Chen and Kuan Cao and Liang Peng and Lin Qu and Minggang Wu and Peng Wang and Shuting Yu and Tingkun Wen and Wensen Feng and Xiaoxiao Xu and Yi Wang and Yichang Zhang and Yongqiang Zhu and Yujia Wu and Yuxuan Cai and Zenan Liu},
      year={2025},
      eprint={2508.02324},
      archivePrefix={arXiv},
      primaryClass={cs.CV},
      url={https://arxiv.org/abs/2508.02324}, 
}

@misc{google2026nanobanana2,
  author       = {{Google DeepMind}},
  title        = {{Nano Banana 2}: Combining Pro capabilities with lightning-fast speed},
  year         = {2026},
  howpublished = {\url{https://blog.google/innovation-and-ai/technology/ai/nano-banana-2/}},
  note         = {Accessed: 2026-06-02}
}

@article{Andersson2020FLIP,
  author    = {Pontus Andersson and
               Jim Nilsson and
               Tomas Akenine{-}M{\"{o}}ller and
               Magnus Oskarsson and
               Kalle {\AA}str{\"{o}}m and
               Mark D. Fairchild},
  title     = "{{FLIP:} {A} Difference Evaluator for Alternating Images}",
  journal   = {Proceedings of the ACM on Computer Graphics and Interactive Techniques},
  volume    = {3},
  number    = {2},
  pages     = {15:1--15:23},
  year      = {2020},
  doi={10.1145/3406183}
}

@inproceedings{lu2025intrinsiccontrolnet,
  author       = {Jiayuan Lu and
                  Rengan Xie and
                  Zixuan Xie and
                  Zhizhen Wu and
                  Dianbing Xi and
                  Qi Ye and
                  Rui Wang and
                  Hujun Bao and
                  Yuchi Huo},
  title        = {IntrinsicControlNet: Cross-Distribution Image Generation with Real
                  and Unreal},
  booktitle    = {{IEEE/CVF} International Conference on Computer Vision, {ICCV} 2025,
                  Honolulu, HI, USA, October 19-25, 2025},
  pages        = {27315--27325},
  publisher    = {{IEEE}},
  year         = {2025},
  url          = {https://doi.org/10.1109/ICCV51701.2025.02536},
  doi          = {10.1109/ICCV51701.2025.02536},
  bibsource    = {dblp computer science bibliography, https://dblp.org}
}


\end{document}


\maketitle


In this supplemental document, we provide details about the architecture of our residual encoder and training details. We also provide more qualitative evaluations on text-guided editing, material editing (normal, albedo, roughness), insertion, removal, relighting and ablations on the adversarial probe capacity and the number of residual tokens. 

\Paragraph{Residual encoder architecture}
Our residual encoder comprises a DINOv2-Large~\cite{oquab2024dinov2} backbone and an attention-based fusion module that maps DINOv2 tokens to a residual embedding. Concretely, we introduce 333 learnable tokens with the same dimensionality as the DINOv2 backbone. These tokens act as queries in a cross-attention module, while the DINOv2 tokens provide keys and values. We additionally add a learned embedding indicating the intrinsic channel type, specifying which residual is being inferred. The resulting 333 tokens are then processed by a self-attention block, followed by layer normalization, and finally projected to 4096 dimensions to match the SD3 text-embedding space.

\Paragraph{Adversarial module architecture}
We also employ an adversarial module during training of the residual encoder to support disentanglement. The module first projects the residual tokens to 1024 dimensions. These tokens are then used as keys/values in a cross-attention module with 256 learned query tokens. The resulting 256 tokens are reshaped into a 
16x16 grid and passed through transposed-convolution layers to produce a 64x64x16 tensor, matching the spatial dimensions and channel count of the VAE-encoded intrinsic representation.

\Paragraph{Training details}
We first pre-train the encoder for 40k steps using the standard diffusion loss. Next, we freeze the encoder and train the adversarial module for 15k steps. Finally, we unfreeze both models and jointly train them for 400k steps.

\begin{figure}[!h]
    \centering
    \includegraphics[width=\linewidth]{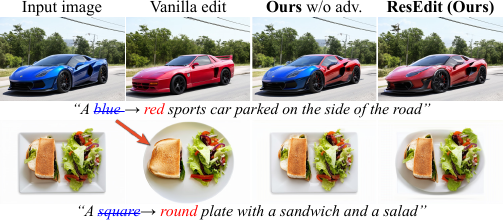}
    \caption{\textbf{Additional text-guided editing results.} From an input image (left), performing diffusion inversion followed by simply replacing the prompt with the edit leads to drastic content change (2nd col.). Our residual anchors the model to the original image (3rd col.), and adversarial optimization is crucial to avoid the leaking of edited attributes into the residual (right).}
    \label{fig:supp_text_based}
\end{figure}

\begin{figure*}
    \centering
    \includegraphics[width=\linewidth]{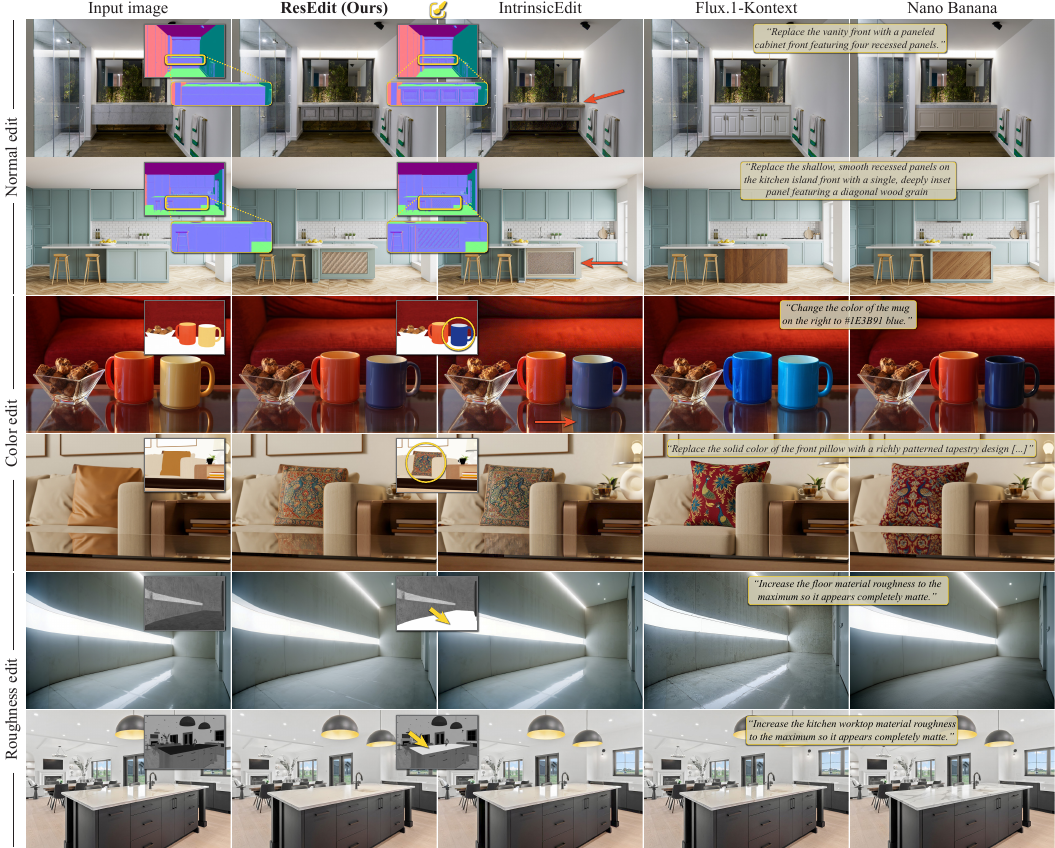}
    \caption{\textbf{Additional material editing results.} We compare our method against IntrinsicEdit, Flux.1-Kontext and Nano Banana on a variety of edits. 
            Best viewed zoomed-in.}
    \label{fig:supp_old_models_material_editing}
\end{figure*}

\begin{figure*}
    \centering
    \includegraphics[width=\linewidth]{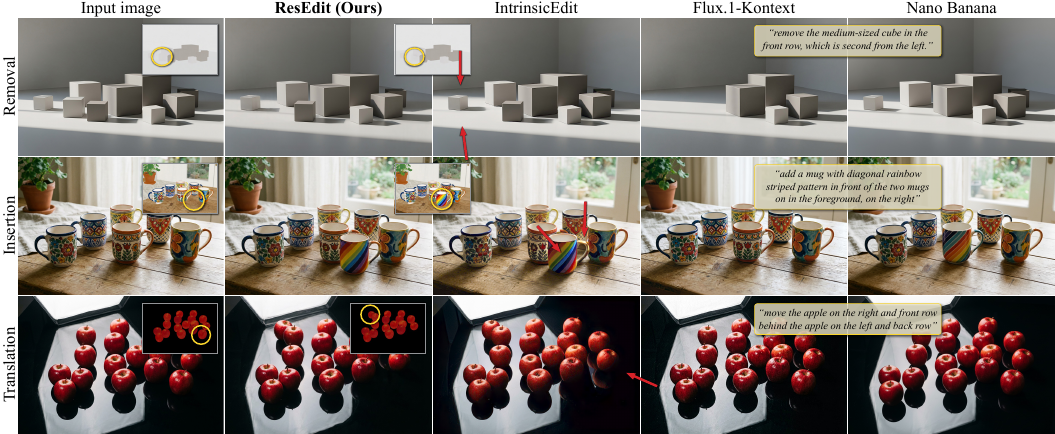}
    \caption{\textbf{Additional insertion and removal results.} We compare our method against IntrinsicEdit, Flux.1-Kontext and Nano Banana on object insertion and removal. 
            Best viewed zoomed-in.}
    \label{fig:supp_old_models_material_editing}
\end{figure*}

\begin{figure*}
    \centering
    \includegraphics[width=\linewidth]{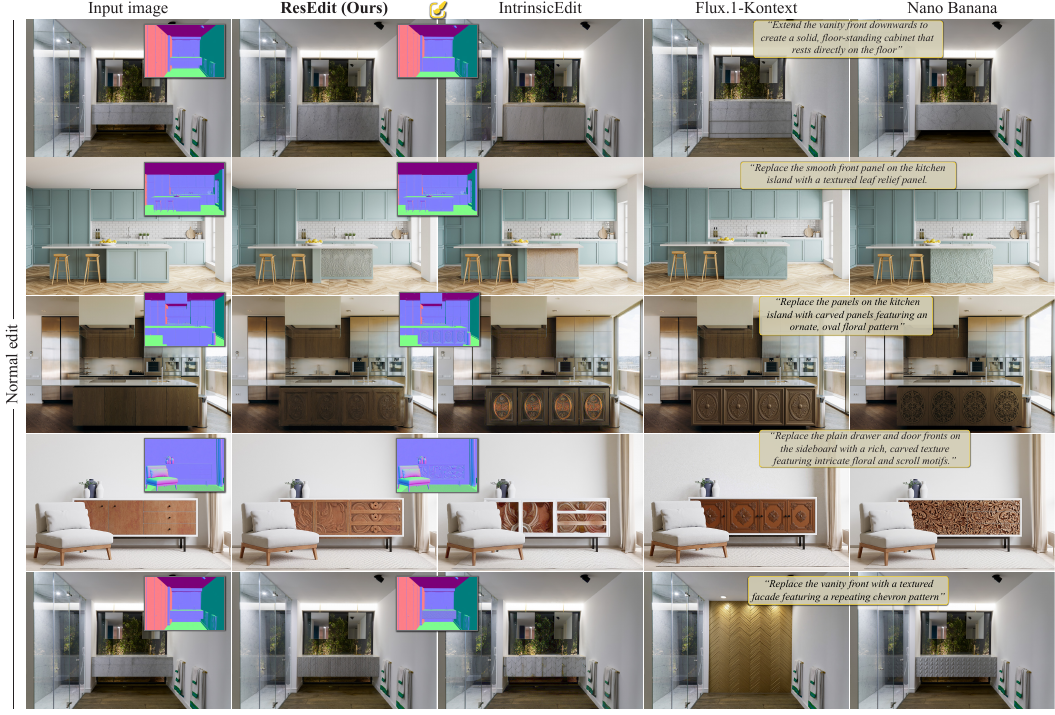}
    \caption{\textbf{Additional normal editing results.} We showcase diverse surface appearance edits on normal maps. Our method (2nd col.) produces more plausible results that better align with the user edit than IntrinsicEdit (3rd col.). Text-based methods (right) struggle to produce precise edits due to the limitations of natural language. 
            Best viewed zoomed-in.}
    \label{fig:supp_normal_editing}
\end{figure*}

\begin{figure*}
    \centering
    \includegraphics[width=\linewidth]{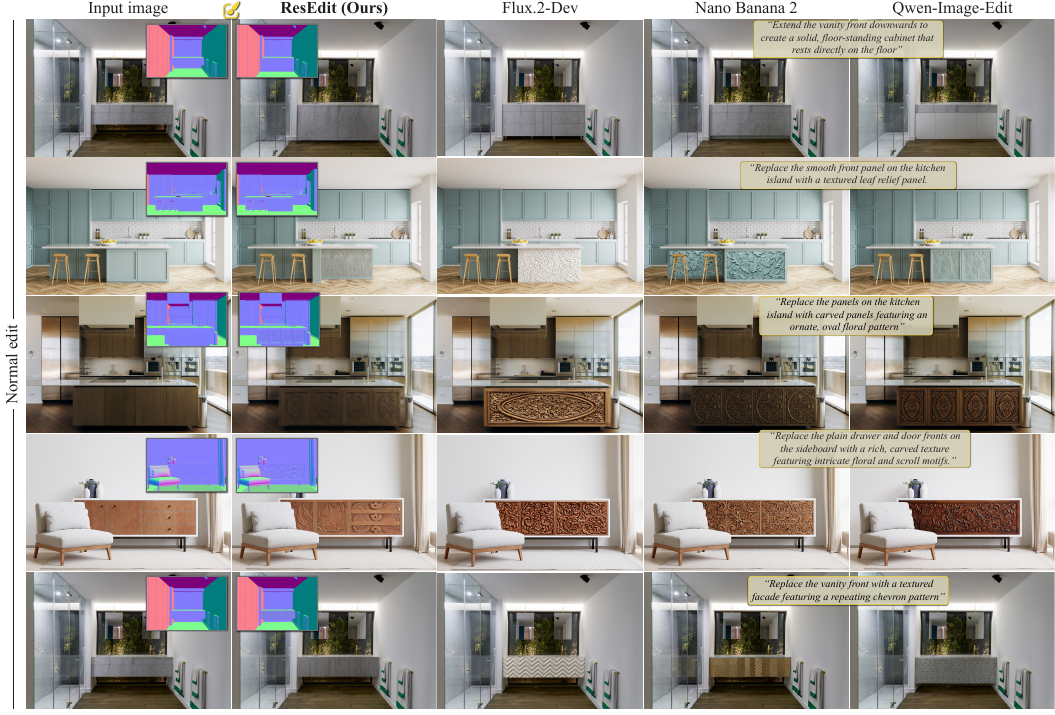}
    \caption{\textbf{Additional normal editing results compared to the recent editing methods.} We showcase diverse surface appearance edits on normal maps. Our method (2nd col.) produces more plausible results that better align with the user edit. Text-based methods (right) struggle to produce precise edits due to the limitations of natural language. 
            Best viewed zoomed-in.}
    \label{fig:supp_normal_editing_new_models}
\end{figure*}

\begin{figure*}
    \centering
    \includegraphics[width=\linewidth]{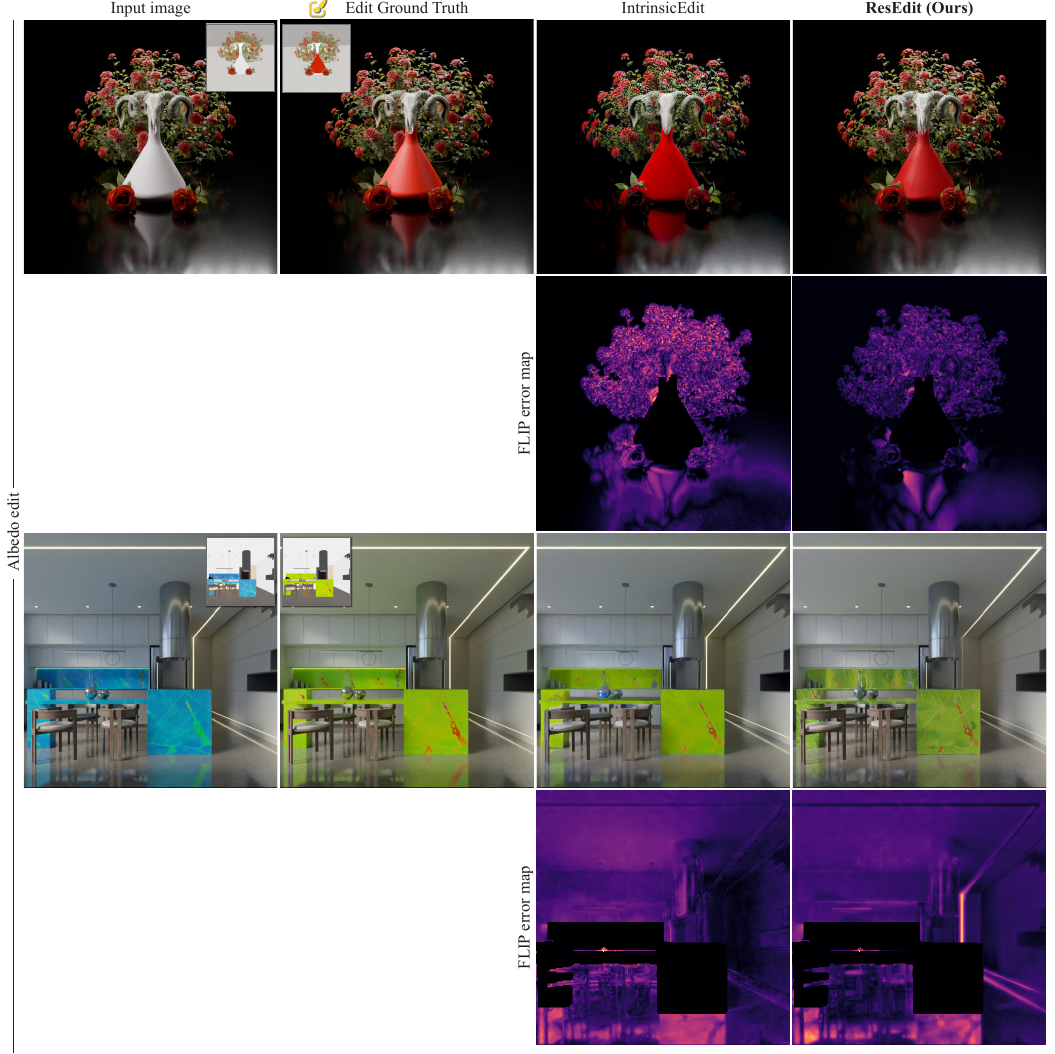}
    \caption{\textbf{Albedo edits on a synthetic dataset.} Our method achieves higher quality results than IntrinsicEdit by preserving the unchanged details and applying edits that are aligned with the edited intrinsics. Best viewed zoomed-in. FLIP error maps are shown only outside the edited regions, using masks computed from the intrinsic-map changes.}
    \label{fig:supp_material_2}
\end{figure*}

\begin{figure*}
    \centering
    \includegraphics[width=\linewidth]{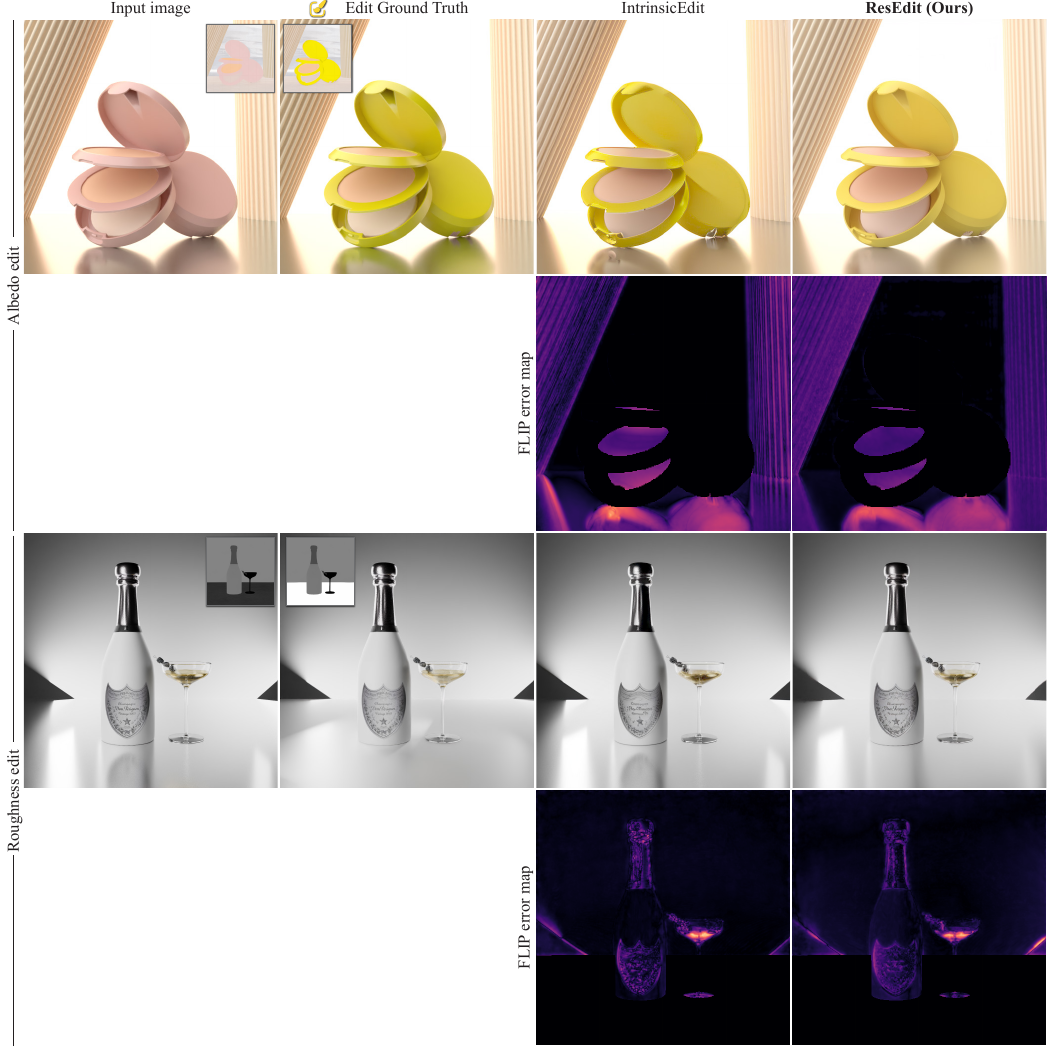}
    \caption{\textbf{Albedo and roughness edits on a synthetic dataset.} Our method achieves higher quality results than IntrinsicEdit by preserving the unchanged details and applying edits that are aligned with the edited intrinsics. FLIP error maps are shown only outside the edited regions, using masks computed from the intrinsic-map changes. Best viewed zoomed-in.}
    \label{fig:supp_material_2}
\end{figure*}

\begin{figure*}
    \centering
    \includegraphics[width=\linewidth]{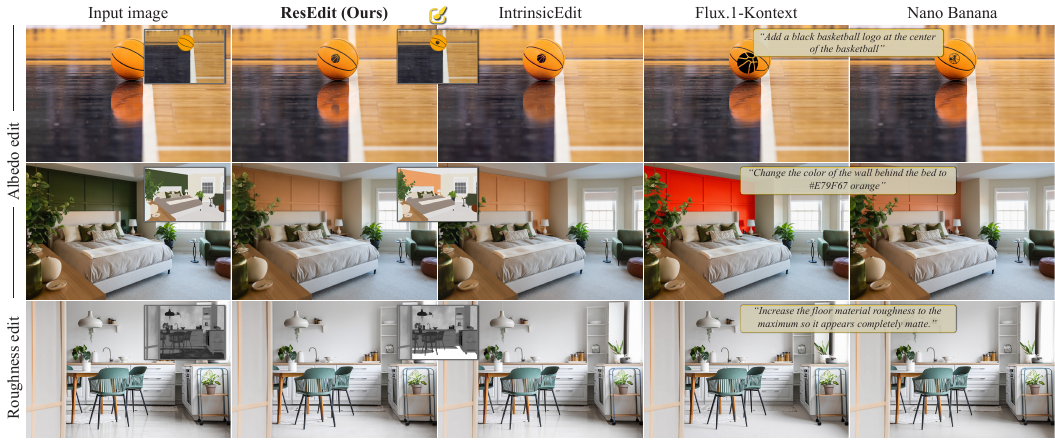}
    \caption{\textbf{Additional albedo and roughness editing results.} We showcase diverse appearance edits on albedo and roughness maps. Our method (2nd col.) produces more plausible results that better align with the user edit than IntrinsicEdit (3rd col.). Text-based methods (right) struggle to produce precise edits due to the limitations of natural language. 
            Best viewed zoomed-in.}
    \label{fig:supp_material_1}
\end{figure*}

\begin{figure*}
    \centering
    \includegraphics[width=\linewidth]{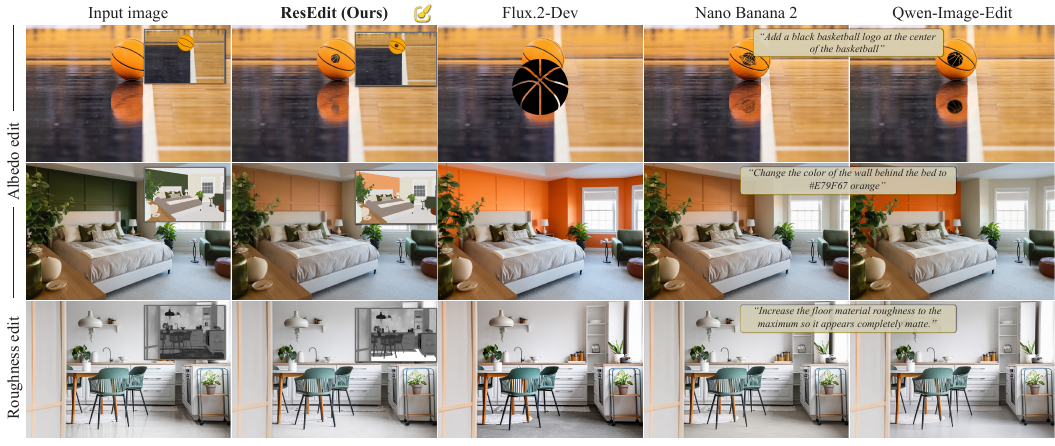}
    \caption{\textbf{Additional albedo and roughness editing results compared to the recent editing models.} We showcase diverse appearance edits on albedo and roughness maps. Our method (2nd col.) produces more plausible and precise results compared to the text-based methods. 
            Best viewed zoomed-in.}
    \label{fig:supp_material_1_new_models}
\end{figure*}

\begin{figure*}
    \centering
    \includegraphics[width=\linewidth]{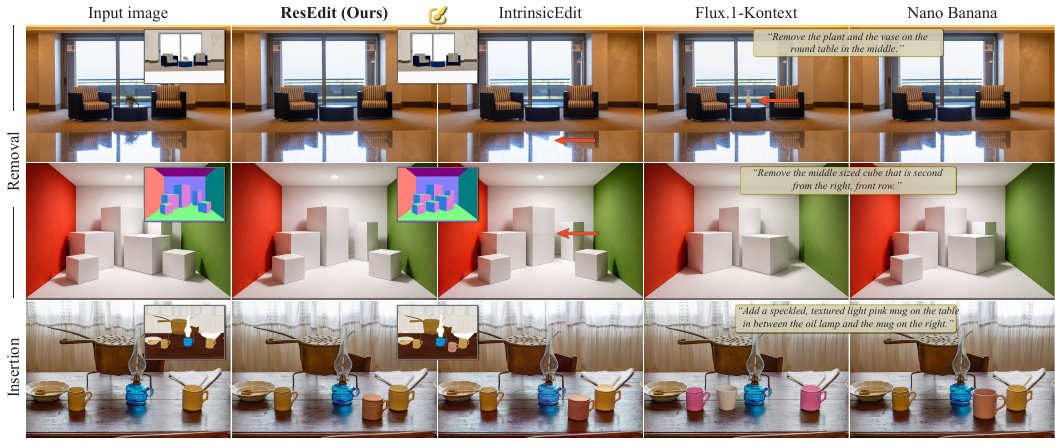}
    \caption{\textbf{Additional insertion and removal results.} Compared to IntrinsicEdit, our method faithfully reconstructs unedited areas while resynthesizing plausible details in edited regions (e.g., shadows and reflections). Flux.1-Kontext and Nano Banana (rightmost columns), shows that text-based editing does not allow for precise editing.
            Best viewed zoomed-in.}
    \label{fig:supp_removal}
\end{figure*}

\begin{figure*}
    \centering
    \includegraphics[width=\linewidth]{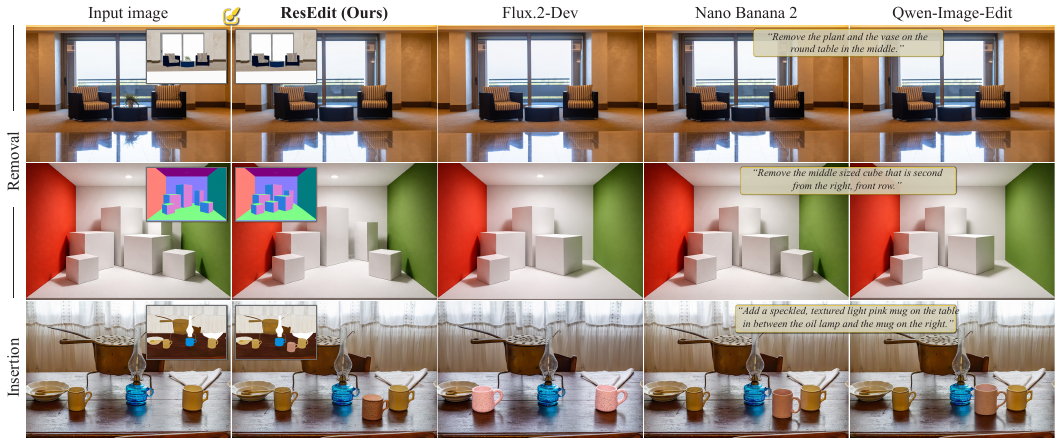}
    \caption{\textbf{Additional insertion and removal results compared to the recent editing methods.} Our method faithfully reconstructs unedited areas while resynthesizing plausible details in edited regions. Text-based methods (right) struggle to produce precise edits due to the limitations of natural language. 
            Best viewed zoomed-in.}
    \label{fig:supp_removal_new_models}
\end{figure*}

\begin{figure*}
    \centering
    \includegraphics[width=\linewidth]{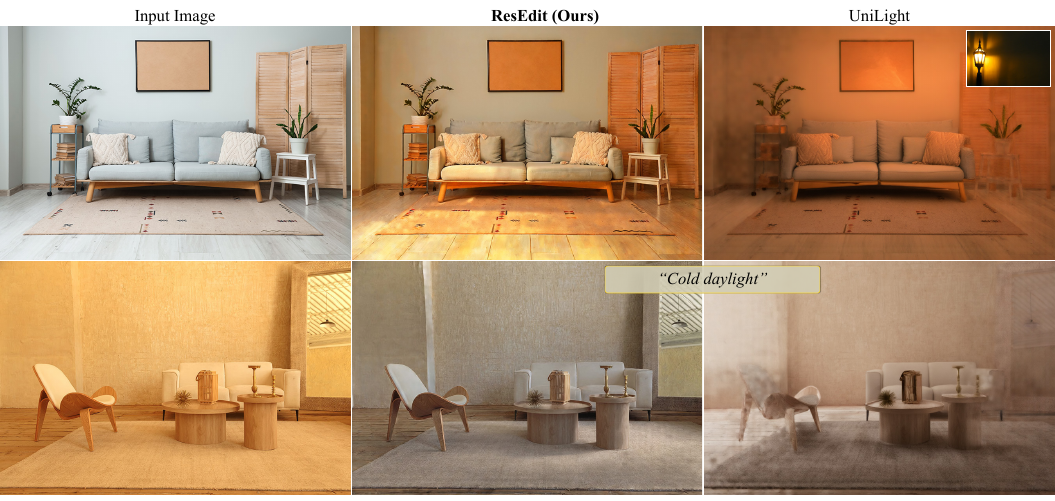}
    \caption{\textbf{Additional relighting results.} Given a lighting description---in the form of text prompts or a reference illumination---we encode it into the UniLight latent space \cite{zhang2025unilight}. We then use the resulting lighting tokens as input conditions for our method, achieving plausible, realistic relighting. The vanilla UniLight approach struggles with identity drifts as it is based on a simple \rgbxrgb pipeline which relies solely on intrinsic channels for identity preservation.}
    \label{fig:supp_relighting}
\end{figure*}

\begin{figure*}
    \centering
    \includegraphics[width=\linewidth]{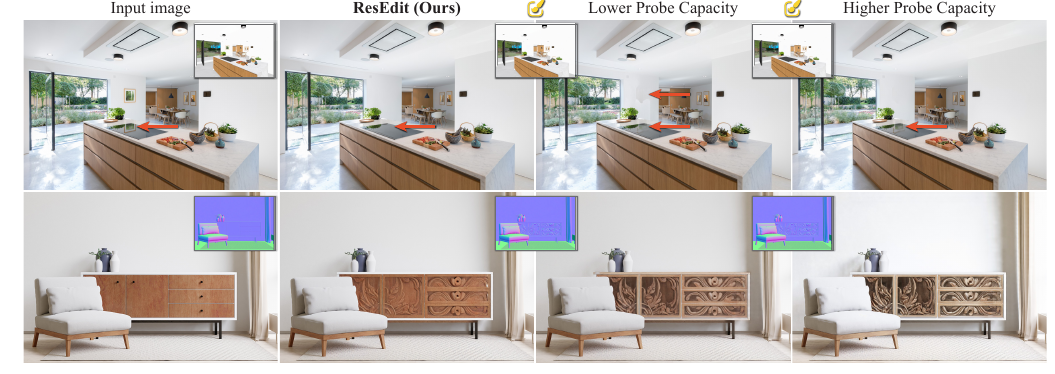}
    \caption{\textbf{Probe capacity ablations.} We compare our default adversarial probe with lower- and higher-capacity variants using one and three hidden layers, respectively. The lower-capacity probe often provides an insufficient disentanglement signal, allowing condition information to leak into the residual and causing editing artifacts. The higher-capacity probe can make the adversarial optimization less stable, leading to oversaturation and artifacts. Our default two-layer probe provides a more reliable balance.}
    \label{fig:supp_probe_capacity}
\end{figure*}

\begin{figure*}
    \centering
    \includegraphics[width=\linewidth]{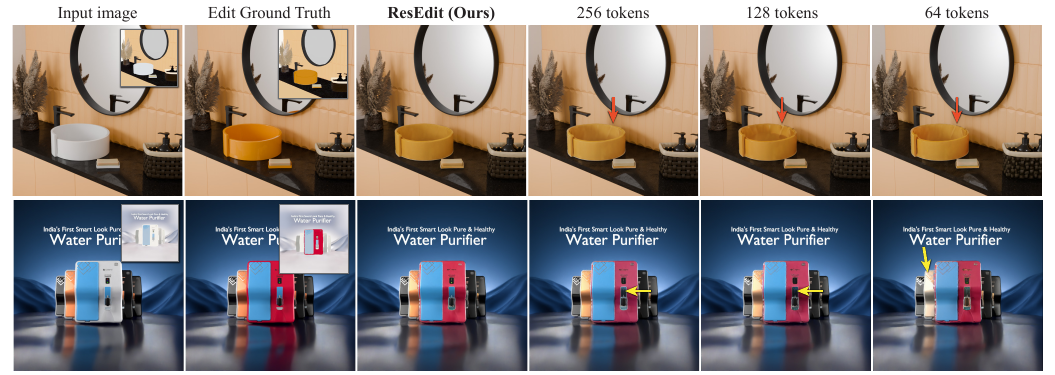}
    \caption{\textbf{Residual token-count ablations.} We compare our default 333-token residual with lower-capacity variants using 256, 128, and 64 tokens. Reducing the number of residual tokens limits the amount of scene-specific information that can be represented, causing fine details and identity cues to be reconstructed less accurately. This reduced residual capacity leads to lower-fidelity edits and more visible artifacts, while the full 333-token residual provides the most reliable reconstruction and editing quality.}
    \label{fig:supp_token_count}
\end{figure*}

\begin{figure*}
    \centering
    \includegraphics[width=\linewidth]{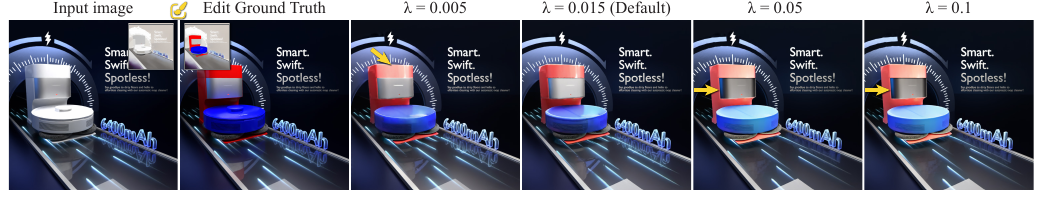}
    \caption{\textbf{Adversarial-weight sensitivity.} We vary the adversarial weight $\lambda$ while keeping all other optimization settings fixed. Small values provide a weak disentanglement signal and can allow condition information to leak into the residual, while large values can destabilize the adversarial optimization and reduce fidelity. Our default $\lambda=0.015$ provides a reliable balance between editability and fidelity. }
    \label{fig:lambda_sensitivity}
\end{figure*}

\bibliographystyle{eg-alpha-doi}
\bibliography{references}